\documentclass[11pt]{article}

\usepackage[final]{acl}
\usepackage{mathtools}
\usepackage{tcolorbox}
\usepackage{times}
\usepackage{latexsym}
\usepackage{pgfplots}
\usepgfplotslibrary{groupplots}
\usepackage{subcaption}
\usepackage[utf8]{inputenc}
\usepackage[ruled,vlined,linesnumbered]{algorithm2e}
\usepackage{hyperref}
\usepackage{amsmath,amssymb}
\usepackage{bbm}
\usepackage{pifont}
\usepackage{enumitem}
\usepackage{booktabs}
\usepackage{pgfplots}
\pgfplotsset{compat=1.17}
\usetikzlibrary{backgrounds}
\usepackage{amsmath}
\usepackage{amssymb}
\usepackage{graphicx}
\usepackage{caption}
\usepackage{multirow}
\usepackage{mathrsfs}
\usepackage{pgfplotstable}
\usepackage{caption}
\usepackage{tikz}
\usepackage{fontawesome5}
\usetikzlibrary{positioning, shapes}
\usetikzlibrary{positioning, shadows}
\usepackage{tabularx}    
\usepackage{array}       
\usepackage{geometry}    
\usepackage{multicol}    
\usepackage{tikz}
\usepackage{xcolor}
\usepackage{siunitx}
\usepackage{threeparttable}
\usepackage{fontawesome5}
\usepackage{subcaption}  
\definecolor{stereoBlue}{HTML}{1F4E79}
\definecolor{stereoOrange}{HTML}{D97706}
\definecolor{stereoGreen}{HTML}{15803D}
\definecolor{stereoRed}{HTML}{B91C1C}
\definecolor{bgGray}{HTML}{F8FAFC}
\definecolor{colSocial} {HTML}{15803D}
\definecolor{colRandom} {HTML}{6B7280}
\definecolor{colNovel}  {HTML}{D97706}
\definecolor{colControl}{HTML}{9CA3AF}
\definecolor{bgGray}    {HTML}{F8FAFC}
\definecolor{colOther}{HTML}{9CA3AF}
\pgfplotsset{
  strip/.style={
    width=4.8cm, height=3.6cm,
    xmin=-1.1, xmax=1.35,
    ymin=0.0,  ymax=3.2,
    ytick={1,2},
    yticklabels={,},
    yticklabel style={font=\scriptsize},
    xtick={-1,-0.5,0,0.5,1},
    xticklabel style={font=\scriptsize,
        /pgf/number format/fixed,
        /pgf/number format/precision=1},
    xlabel={Projection score},
    xlabel style={font=\scriptsize},
    axis background/.style={fill=bgGray},
    grid=major, grid style={line width=0.3pt, draw=white},
    clip=false,
  },
}

\usetikzlibrary{positioning}
\usepackage[T1]{fontenc}

\usepackage[utf8]{inputenc}

\usepackage{microtype}

\usepackage{inconsolata}
\usepackage{xspace}
\usepackage{graphicx}
\newcommand{\ProbingDataset}{\mathcal{F}}

\newcommand{\concepts}{\mathcal{C}}
\newcommand{\randomconcepts}{\mathcal{C'}}
\newcommand{\axis}[2]{\texttt{#1}$\leftrightarrow$\texttt{#2}} 

\newcommand{\vs}[2]{\textit{#1} vs.\ \textit{#2}}
\newcommand{\ours}{\textsc{StereoDisco}\xspace}
\newcommand{\llama}{\textsc{Llama-3-8B-Instruct}\xspace}
\newcommand{\mistral}{\textsc{Mistral-7B-Instruct}\xspace}

\newcommand{\dmodel}{d_{\mathrm{model}}}
\newcommand{\dhead}{d_{head}}
\newcommand{\secref}[1]{\S\kern-0.05em\ref{#1}}
\newcommand{\phead}[1]{\noindent\textbf{#1.}}

\author{
  \textbf{Farane Jalali Farahani\textsuperscript{1}},
  \textbf{Corina Dima\textsuperscript{1}},
  \textbf{Mojtaba Nayyeri\textsuperscript{1}},
  \textbf{Raphael H. Heiberger\textsuperscript{2}},
  \textbf{Steffen Staab\textsuperscript{1,3}}
\\
\\
  \textsuperscript{1}Institute for Artificial Intelligence, Stuttgart, Germany \\
  \textsuperscript{2}Institute for Social Science, Stuttgart, Germany \\
  \textsuperscript{3}University of Southampton, Southampton, United Kingdom
\\
\\
  \small{
    \textbf{Correspondence:} 
    \href{mailto:farane.jalali-farahani@ki.uni-stuttgart.de}{farane.jalali-farahani@ki.uni-stuttgart.de}
  }
}
  
\begin{document}


\newcommand{\steffen}[1]{\textcolor{red}{Steffen: #1}}
\newcommand{\farane}[1]{\textcolor{blue}{Farane: #1}}
\newcommand{\corina}[1]{\textcolor{teal}{(Corina: #1)}}
\newcommand{\mojtaba}[1]{\textcolor{green}{Mojtaba: #1}}

\title{\ours: Discovering Stereotypicality in LLMs }

\maketitle
\begin{abstract}
LLMs encode, convey, and perpetuate stereotypes. Prior computational research focuses on a small set of
semantic axes investigated in social psychology, and operates on
word embeddings produced by
language models, leaving open which other semantic axes carry stereotypical associations in LLMs and how LLMs internally represent such axes. We introduce
\ours, a framework that adapts the  \textit{semantic differential} method \citep{osgood1957semantic}
 to the systematic study of stereotypes in LLM 
internal representations. 
\ours constructs
$\sim$2{,}000 candidate \textit{semantic axes} from WordNet antonym synsets,
recovers each as a \textit{geometric axis} in the LLM's activation space via
probing, and identifies stereotypical axes via a
statistical test over concept projections.
As a case study, we apply \ours to social group stereotypes with \llama and \mistral. We find that the two LLMs agree with each other on
social group ratings more than with humans, suggesting that LLM-encoded  stereotype content diverges from that documented in social psychology. We also discover
stereotypical axes not investigated in prior 
work -- including \vs{humble}{proud},
\vs{narrow-minded}{broad-minded}, and \vs{cowardly}{brave}, which
human annotators independently confirm.

\textcolor{red}{\textbf{Warning:} This paper contains content that
some readers may find offensive.}
\end{abstract}

\section{Introduction}

The \textit{semantic differential} method \citep{osgood1957semantic} is an established psychometric method for
constructing a \textit{semantic axis} from a pair of opposites and then rating concepts along it. For example, one can construct a semantic axis from opposite adjectives such as \textit{inferior} and \textit{superior} and then rate where concepts such as ``CEOs'', ``employees'', and ``interns'' would be placed along the axis. The method can be applied to any pair of opposites, e.g. \vs{love}{hate}, \vs{to like}{to dislike}, and any concept of interest: food items, brands, individuals, social groups, etc.

The semantic differential method has been recently operationalized in computational research  using word embeddings \citep{an-etal-2018-semaxis,kozlowski2019geometry,lucy-etal-2022-discovering, adaptive_axes}, which map both semantic axes and concepts into a shared vector space and measure their association via cosine similarity.  This line of work, however, relies on word embeddings as outputs of language models, leaving open the question of how the semantic differential method can be applied to the internal representations of large language models (LLMs).

In this paper, we answer this question by redesigning the semantic differential method for the study of internal representations of LLMs.  We hypothesize that \textbf{(H1)} LLMs encode each semantic axis as a \textit{geometric axis} --- a direction in their activation space; \textbf{(H2)} \textit{rating a concept} on the semantic axis corresponds to \textit{projecting the concept} along the corresponding geometric axis; \textbf{(H3)} along a stereotypical semantic axis, projections of relevant concepts differ significantly from projections of random concepts. 

To test these hypotheses, we introduce \ours, a framework with the following components: (i) a \textbf{semantic axis source component}, which supplies candidate semantic axes as pairs of opposites; (ii) a \textbf{semantic axis representation component}, which maps each semantic axis to a geometric axis in the model's internal space; (iii) a \textbf{concept projection component}, which locates concept mentions along the recovered geometric axes; and (iv) a \textbf{stereotypicality test
component}, which decides whether projections of relevant concepts differ from those of a reference set.
\ours is applicable to any concept, including social groups \citep{davani}, food and beverages \citep{kozlowski2019geometry}, brands \citep{brands_stereotypes} -- wherever a meaningful set of opposites exists. 

As a case study, we apply \ours to \textit{social group stereotypes} --- ``mental associations between the name of a
social group (e.g., accountant) and various attributes (e.g., nerdy, boring, intelligent)'' \citep[p.~374]{stereotype_role}. This definition lends itself naturally to the semantic differential method: a social group (e.g., ``CEOs'') can be located along a semantic axis defined by opposing attributes (e.g.,
\vs{inferior}{superior}), and its position indicates the strength
of the stereotypical association with each pole. LLMs are known to encode such associations
\citep{NicolasCaliskan2025ChatbotStereotypes}, and given their
increasing use in domains such as hiring
\citep{nghiem-etal-2024-gotta} and healthcare
\citep{Zack2024GPT4BiasHealthcare}, these associations may affect
downstream decisions. While H1-H3 concern the framework, we test a further hypothesis specific to this case study: \textbf{(H4)} social group stereotypes encoded by LLMs align with human stereotypes.

To test H4, we compare \ours to human judgments along two
aspects. \textit{First}, do \ours's projections of social
groups onto geometric axes by \ours align with human ratings on the corresponding semantic axes? This
measures agreement on the \textit{content} of a stereotype, e.g.,
whether \ours places ``CEOs'' on the same side of the
\vs{inferior}{superior} axis as humans do.
\textit{Second}, do the semantic axes which \ours identifies as stereotypical align with those identified by humans? This measures
agreement on \textit{which} semantic axes carry stereotypical associations.
We answer the questions using existing psychological surveys and  human annotations we collect.

\paragraph{Contributions.} Our main contributions are:
\begin{itemize}[leftmargin=*,itemsep=2pt,topsep=2pt]
    \item \textbf{A theory-grounded framework.} \ours
          quantifies the stereotypicality of semantic axes in LLM
          internal representations via four components (semantic axis source, semantic axis representation, concept projection, stereotypicality test), applicable to any concept family with a meaningful set of opposites.
    \item \textbf{Internal localization.} Activation
          probing across all layer-head pairs of \llama and
          \mistral shows that semantic axes are linearly encoded as
          geometric axes in a small subset of middle-to-late attention heads.
    \item \textbf{Case study on social group stereotypes.} Applying \ours to social group stereotypes with \llama and
          \mistral, we find that the
          two LLMs agree on social group ratings more with each other than with humans. We also discover stereotypical semantic axes not investigated in
          prior work, including \vs{humble}{proud},
          \vs{narrow-minded}{broad-minded}, and
          \vs{cowardly}{brave} - which human annotators independently confirm.
\end{itemize}

\noindent Overall, \ours moves the study of stereotypes in
LLMs from the analysis of word embeddings to the models' internal representations, while at the same time enabling explorations that go
beyond the small, predefined set of semantic axes investigated in
social psychology.

\section{Background}

\subsection{WordNet}
\label {sec:wordnet}
WordNet \cite{miller-1992-wordnet} organizes English words into \textit{synsets} --- sets of synonyms sharing the same sense. Each \textit{sense} is one aspect of a word's meaning, paired with a  \textit{definition} describing it \cite{jurafsky-martin-2020}. We refer to synsets using the WordNet notation \texttt{lemma.pos.index}, where \texttt{lemma} is the word form, \texttt{pos} is the part-of-speech tag (e.g., \textit{a} for adjective, \textit{n} for noun), and \texttt{index} is the sense number. For example, the synset \textit{large.a.01} groups synonyms such as \textit{large} and \textit{big}, and is defined as ``above average in size, number, quantity, magnitude, or extent''.
Importantly, 
WordNet defines antonymy at the sense level rather than at the word level: \textit{large.a.01} is antonymous 
to \textit{small.a.01}.

\subsection{Decoder-only LLMs and Probing}
\label{sec:llm_and_probing}
Decoder-only LLMs are based on transformers \cite{vaswani} that map an input token sequence to hidden representations iteratively refined across $L$ layers, each containing $H$ attention heads and an MLP block. For a given input token $e$, the embedding layer produces an initial hidden representation $r_{0} = \mathrm{Embed}(e) \in \mathbb{R}^{\dmodel}$, where $\dmodel$ is the model's hidden dimension. At each layer $\ell$, each attention head $h$ projects $r_{\ell-1} \in \mathbb{R}^{\dmodel}$ into a lower-dimensional subspace of dimension $\dhead$ (the per-head dimension, with $\dhead < \dmodel$) via the projection matrix $O_{\ell,h} \in \mathbb{R}^{\dhead \times \dmodel}$, and computes 
$x_{\ell,h} = \mathrm{ATTN}_{\ell,h}(O_{\ell,h}\, r_{\ell-1}) \in \mathbb{R}^{\dhead}$, 
where $\mathrm{ATTN}_{\ell,h}(\cdot)$ is the attention head $h$ from layer $\ell$.





\noindent\textit{Probing}~\cite{belinkov-2022-probing} is a method for investigating which features are
encoded in an LLM's internal representations by using a
classifier --  a \textit{probe} -- to predict a target concept
from the LLM's representations.
Probing has been applied to decoder-only LLMs to identify which layers and attention heads encode high-level  concepts such as \textit{truthfulness} (e.g. "factual" vs.\ "non-factual") \cite{marks2024geometrytruthemergentliner} or \textit{political perspective} (e.g. "liberal" vs.\ "conservative") \cite{kim2025linear}. Following this work, 
we refer to $x_{\ell,h}$ as the \textit{activation} of attention head $h$ 
in layer $\ell$. During probing, the inputs (prompts) associated with ground truth labels for a concept of interest are passed through the LLM, and the resulting activations are recorded;
a \textit{linear probe} is then fit to these activations to test whether the concept is \emph{linearly encoded} in the representations of the LLM -- a property useful for \textit{steering} the LLM outputs via activation interventions \cite{li_intervention, hernandez2024linearity}.

\section{Related Work}

\subsection{Psychological Models of Stereotype}
\label{sec:psy-model}


The \textit{Stereotype Content Model} \citep[SCM;][]{fiske2002model, fiske2006universal, fiske2018stereotype} is one of the most influential psychological models of stereotype, organizing stereotypes along two dimensions: \textit{Warmth} -- the intention to help or harm  (e.g.,  thoughtful, unfriendly) -- and \textit{Competence} -- the ability to carry out that intention (e.g., skilled, ineffective). Recent models have proposed extensions of these SCM categories: \citet{abele} subdivide Warmth into \textit{Sociability} and \textit{Morality}, and Competence into \textit{Ability} and \textit{Assertiveness (Agency)}. \citet{koch2016abc} extend the SCM with \textit{beliefs} (religious-secular beliefs and political orientation) and \textit{status} (perceived social and economic standing).
Building on these models, \citet{nicolas_2021} compiled a \textit{stereotype dictionary} 
of adjectives and nouns organized into seven categories: 
Sociability, Morality, Ability, Agency, 
Status, Politics, and Religion -- which we use in our experiments.

\subsection{Computational Models of Stereotype}
Computational work on stereotypes in text and LLMs has largely followed two paradigms.

\noindent \textbf{The first paradigm} builds on the Word Embedding Association Test \citep[WEAT;][]{caliskan2017semantics}, which adapts the Implicit Association Test from social psychology and quantifies bias through cosine similarity between embeddings of target and attribute words. Subsequent work extended this paradigm to contextualized representations through the Contextualized Embedding Association Test \citep[CEAT;][]{ceat} and the Sentence Encoder Association Test \citep[SEAT;][]{seat}.
Stereotypes in social psychology, however, are defined as positions of social groups along axes constructed from bipolar attributes~\citep{stereotype_role, fiske2002model}, calling for a method that locates concepts along such axes rather than measuring associative strength between sets.

\noindent \textbf{The second paradigm} builds on the semantic differential method, in which a semantic axis is constructed from a pair of opposites and concepts are rated on the axis. Prior work in this paradigm has 
quantified stereotypes in news \citep{adaptive_axes}, 
Wikipedia \citep{lucy-etal-2022-discovering}, social media  \citep{fraser2022computational},  
and LLMs~\cite{schuster2025profiling}. 
Many of these studies build on the \textit{POLAR} framework~\cite{Mathewpolar} and its sense-aware extension \textit{SensePOLAR}~\cite{engler-etal2022sensepolar}, in which each semantic axis is a direction in the word embedding space, given by the difference $\mathbf{a}_i = \mathbf{w}^{+}_{i} - \mathbf{w}^{-}_{i}$ between the embeddings of two opposites. A word $v$ with embedding $\mathbf{w}_v$ is then projected onto this direction as $w'_{v,i} = a_i^\top w_v$.

\phead{Limitations of existing paradigms} Existing paradigms have two limitations. \textit{First}, they operate on a small, predefined 
set of theory-driven semantic axes investigated in social 
psychology -- for example, the stereotype dictionary of \citet{nicolas_2021}, used in recent computational work~\citep{fraser-etal-2021-understanding, fraser2022computational, omrani, schuster2025profiling} -- leaving open whether LLMs encode stereotypical associations along semantic axes beyond this dictionary. \citet{NicolasCaliskan2025ChatbotStereotypes} instead prompt LLMs to generate associations about social groups and apply cluster analysis to identify stereotypical content; we build a large-scale probing dataset and systematically evaluate semantic axes inside LLM representations.
\textit{Second}, existing paradigms use word embeddings or 
layer-averaged LLM representations rather than examining the internal representations of LLMs. \citet{scout} investigate stereotypes at the level of attention heads, but focus on controlling LLM outputs for a limited set of predefined stereotypical semantic axes rather than systematically identifying stereotypical semantic axes from a large candidate pool. To address both limitations, we design \ours (\secref{sec:method}).

\section{\ours Framework}
\label{sec:method}
We propose \ours, a framework that discovers semantic axes in LLMs and quantifies their stereotypicality with respect to relevant concepts. The framework has four components, each with explicit design choices that can be instantiated differently:
(i) a \textbf{semantic axis source component}, which supplies candidate axes as pairs of opposites and operationalizes each pair into the labeled inputs required by the next component; (ii) a \textbf{ semantic axis representation component}, which maps each semantic axis to a geometric axis in the LLM's internal space; (iii) a \textbf{concept projection component}, which locates concept mentions along that geometric axis; and (iv) a \textbf{stereotypicality test
component}, which decides whether projections of relevant concepts differ from those of a reference set. We
describe each component below; the choices we make in this paper are one instantiation, other explorations are possible under different instantiations of the framework.


\subsection{Semantic Axis Source Component}
\label{sec:constructing-axis}

A semantic axis can in principle be constructed from any pair of
opposites --- adverbs (\vs{accidentally}{purposely}), proper nouns
(\vs{Kondo}{messy}\footnote{\href{https://en.wikipedia.org/wiki/Marie_Kondo}{Marie Kondo}, the Japanese tidying consultant, used here for a tidy vs.\ messy axis.}, \vs{Warren}{Lee}\footnote{Used by \citet{kim2025linear} for a liberal vs.\ conservative axis.}),  adjectives, etc.
The semantic axis source component supplies the candidate pool of opposite pairs and operationalizes each pair into the labeled inputs the axis representation component requires to construct the two opposing \textbf{poles} of the axis. The choice of source is left to the user: WordNet, ConceptNet \citep{conceptnet}, domain-specific dictionaries, or hand-curated lists are all admissible. In this paper we use adjectives, which most directly describe stereotype attributes, and draw them from WordNet  for two reasons: it provides thousands of antonymous adjective pairs, and it organizes them into sense-disambiguated synsets (\secref{sec:wordnet}). The form of the labeled inputs is dictated by the representation method used in \secref{sec:linear-probing}.

\noindent \textbf{Probing Dataset.} We create a probing dataset $\ProbingDataset$ by retrieving all $M = 1{,}999$ available antonymous-synset pairs from WordNet, comprising $5{,}182$ unique adjectives; each pole is a synset that may contain multiple adjectives (e.g., \textit{large.a.01} includes \textit{large} and \textit{big}). Formally, each semantic axis is a pair $(p^{-}, p^{+})$ of antonymous synsets, and $\mathcal{SA}$ denotes the set of all such pairs. 

Obtaining a geometric axis in the LLM's representation space requires labeled examples for each pole (\secref{sec:llm_and_probing}).  We generate sentence examples by inserting each pole into sentence templates, each containing two placeholders: \{ADJ\}, replaced by an adjective from the pole's synset, and \{DEF\}, replaced by its WordNet definition, which disambiguates the sense. 
We use two template formulations, \textit{Simple} 
(e.g., \textit{``\{ADJ\} means: \{DEF\}.''}) and \textit{Listing} 
(e.g., \textit{``List adjectives similar to \{ADJ\}, in the sense of \{DEF\}.''}), 
each with $30$ templates (Table~\ref{tab:templates}); an ablation study 
motivating the choice of \textit{Listing} templates is reported in 
Appendix~\ref{app:ablation}.
 
Because synsets contain a varying number of synonyms, we instantiate each template once per adjective in the pole's synset, then downsample to $N = 30$ sentences per pole to obtain a balanced sample. The resulting dataset $\ProbingDataset = \bigcup_{j=1}^{M} \left(\ProbingDataset_j^{+} \cup \ProbingDataset_j^{-}\right)$ contains $1999 \times 2 \times 30$ labeled sentences, where $\ProbingDataset_j^{\pm} = {(s_{i,j}^{\pm}, \pm 1)}_{i=1}^{N}$, where $j$ indexes the semantic axis, $i$ indexes the template instantiations, and $s_{i,j}^{+}$ and $s_{i,j}^{-}$ are the $i$-th instantiated sentences for the positive and negative poles of semantic axis $j$, respectively.

\subsection{Semantic Axis Representation Component}
\label{sec:linear-probing}

The semantic axis representation component maps each semantic axis to a geometric axis in the model's internal representation space, given the labeled inputs supplied by \secref{sec:constructing-axis}. Instantiating it requires four choices: (i) \textit{where} in the model to probe; (ii) \textit{from which token position} to extract activations; (iii) how to derive a \textit{geometric axis} from the labeled examples; and (iv) \textit{which} of the resulting candidate geometric axes to keep, since probing at every layer–head pair yields many candidates per axis. We describe each in turn and state our instantiation.

\phead{Probing target} Following prior work on locating high-level concepts inside decoder-only LLMs \cite{marks2024geometrytruthemergentliner, kim2025linear}, we probe the output of each attention head. For a sentence $s$, a layer $\ell$, and head $h$, we denote the head output by $\mathbf{x}_{\ell,h}(s)\in \mathbb{R}^{\dhead}$, where $\dhead$ is the per-head output dimensionality ($\dhead=128$ for \llama and \mistral).
Passing each labeled sentence from \secref{sec:constructing-axis} through the LLM and reading the head output at layer $\ell$ and head $h$ yields a \textit{labeled activation}. A positive-pole sentence
$s^{+}_{i,j} \in \ProbingDataset^{+}_j$ yields the activation $(\mathbf{x}_{\ell, h}(s^{+}_{i,j}),\, +1)$, while a negative-pole sentence $s^{-}_{i,j} \in \ProbingDataset^{-}_j$ yields 
$(\mathbf{x}_{\ell, h}(s^{-}_{i,j}),\, -1)$, with 
$i \in \{1,\ldots,N\}$, $j \in \{1,\ldots,M\}$, 
$\ell \in \{1,\ldots,L\}$, and $h \in \{1,\ldots,H\}$. 

\phead{Token position} Activations must be read from a chosen token position. Two common conventions are the \textit{last
token}~\citep{marks2024geometrytruthemergentliner,
bao-etal-2025-probing} and the \textit{mean over all
tokens}~\citep{palma-etal-2025-llamas}. We use the mean over all tokens: in our setting, the last token concentrates probe signal in very early layers, inconsistent with the established finding that semantic representations emerge in middle-to-late layers of decoder-only LLMs~\citep{marks2024geometrytruthemergentliner,
lavi-etal-2026-detecting, poulis2026testinglimitstruthdirections}, whereas the token mean recovers that pattern. An ablation study related to token position is reported in Appendix~\ref{app:ablation}.


\phead{Probing method} Given labeled activations for axis $j$ at $(\ell, h)$, the probing method maps them to a geometric axis $\boldsymbol{\theta}_{j,\ell,h} \in \mathbb{R}^{\dhead}$.
We use \textit{mass-mean probing}~\cite{li_intervention, marks2024geometrytruthemergentliner} rather than alternatives such as logistic regression \cite{bao-etal-2025-probing, marks2024geometrytruthemergentliner}. Mass-mean yields the geometric axis as a deterministic difference between pole means: 

\begin{equation}
\boldsymbol{\mu}^{+}_{j,\ell,h} = \frac{1}{N}\sum_{i=1}^{N} \mathbf{x}_{\ell,h}(s^{+}_{i,j}),
\end{equation}
\begin{equation}
\boldsymbol{\mu}^{-}_{j,\ell,h} = \frac{1}{N}\sum_{i=1}^{N} \mathbf{x}_{\ell,h}(s^{-}_{i,j}),
\end{equation}
\begin{equation}
\label{eq:theta}
    \boldsymbol{\theta}_{j,\ell,h} = \boldsymbol{\mu}^{+}_{j,\ell,h} - \boldsymbol{\mu}^{-}_{j,\ell,h},
\end{equation}

\noindent so swapping which pole is labeled $+1$ and which is $-1$ simply 
negates $\boldsymbol{\theta}_{j,\ell,h}$, leaving the underlying axis 
unchanged. Logistic regression, in contrast, depends on 
optimization choices and need not preserve the axis under
label swap. 


\phead{Candidate selection} Probing at every layer-head pair yields $L \times H$ candidate geometric axes per semantic axis (1,024 for \llama and \mistral, both with $L = H = 32$). We score each candidate by the \textit{ratio of ``between-class'' to ``within-class'' variance} of the labeled activations projected onto $\boldsymbol{\theta}_{j,\ell,h}$ \cite{burger2024truth, bao-etal-2025-probing, poulis2026testinglimitstruthdirections}:

\begin{equation}
R_{j,\ell,h} = \frac{\|\boldsymbol{\mu}^{+}_{j,\ell,h} - \bar{\boldsymbol{\mu}}_{j,\ell,h}\|^2 + \|\boldsymbol{\mu}^{-}_{j,\ell,h} - \bar{\boldsymbol{\mu}}_{j,\ell,h}\|^2}{\mathrm{Var}(\mathbf{x}_{\ell,h}(s^{+}_{\cdot,j})) + \mathrm{Var}(\mathbf{x}_{\ell,h}(s^{-}_{\cdot,j}))},
\label{eq:score}
\end{equation}

\noindent where $\bar{\boldsymbol{\mu}}_{j,\ell,h}$ is the mean over all positive- and negative-labeled sentences of axis $j$, and 

\begin{equation}
\mathrm{Var}(\mathbf{x}_{\ell,h}(s^{\pm}_{\cdot,j})) = \frac{1}{N}\sum_{i=1}^{N} \bigl\|\mathbf{x}_{\ell,h}(s^{\pm}_{i,j}) - \boldsymbol{\mu}^{\pm}_{j,\ell,h}\bigr\|^2
\end{equation}

\noindent is the total within-class variance of each pole. A higher $R_{j,\ell,h}$ indicates greater linear
separability of the two poles in the head ($\ell, h$)'s activation space, and thus stronger evidence that axis $j$ is \emph{linearly
encoded} there~\cite{burger2024truth, bao-etal-2025-probing, poulis2026testinglimitstruthdirections}. The scored candidates feed into \secref{sec:projection}; how many top-scoring candidates to retain is itself an instantiation choice, fixed there.




\subsection{Concept Projection Component}
\label{sec:projection}
The concept projection component takes the scored geometric axes produced by \secref{sec:linear-probing} and locates each concept mention $c \in \concepts$ along the corresponding semantic axis as a scalar $\pi_{j}(c)$. Instantiating it requires four choices: (i) \textit{how to obtain an activation} for $c$; (ii) \textit{how to project} that activation onto a geometric axis; (iii) \textit{how to rescale} per-head projections; and (iv) \textit{how to aggregate} across the $L \times H$ candidates per axis. We describe each in turn and state our instantiation.

\phead{Concept activation}
For $c \in \concepts$, we construct a prompt $q$ of the form \textit{``Provide the best description of \{c\} using adjectives.''} We feed $q$ to the LLM and extract the activation $\mathbf{x}_{\ell,h}(q) \in \mathbb{R}^{\dhead}$ at layer $\ell$ and head $h$, using the same token-position convention as in \secref{sec:linear-probing} (mean over all tokens). 

\phead{Projection} We project $\mathbf{x}_{\ell,h}(q)$ onto the geometric axis $\boldsymbol{\theta}_{j,\ell,h}$:
\begin{equation}
\pi_{j,\ell,h}(c) =
\frac{\mathbf{x}_{\ell,h}(q)^{\top}\, \boldsymbol{\theta}_{j,\ell,h}}
{\lVert \boldsymbol{\theta}_{j,\ell,h} \rVert_2^{2}}.
\label{eq:projection}
\end{equation}

\phead{Per-head rescaling} The resulting projections are rescaled by z-scoring per head, jointly over $\concepts$ and the reference set $\randomconcepts$ introduced in \secref{sec:hypothesis}. This
puts each head on a common scale (mean 0, unit variance) so projections can be aggregated across heads and compare across semantic axes, while preserving the distributional shift between $\concepts$ and $\randomconcepts$ that the test in \secref{sec:hypothesis} detects. 

\phead{Aggregation}
We select the top-$k$ layer--head pairs with the highest 
variance-ratio scores (Eq.~\ref{eq:score}) and average their 
rescaled projections (Eq.~\ref{eq:agg_projection}). The value of $k$ is an instantiation choice, fixed in \secref{sec:comparison}.
\begin{equation}
\pi_{j}(c) = \frac{1}{k} \sum_{(\ell, h) \in \mathrm{top}\text{-}k} \pi_{j,\ell,h}(c).
\label{eq:agg_projection}
\end{equation}

\subsection{Stereotypicality Test Component}
\label{sec:hypothesis}
The stereotypicality test component takes per-axis projections from \secref{sec:projection} and identifies semantic axes along which stereotypical associations occur. We hypothesize that along a stereotypical axis, the projections of a relevant group of concepts  differ significantly 
from those of a random group of concepts. Instantiating this component requires three choices: (i) a \textit{reference set} against which to compare relevant-concept projections; (ii) a \textit{statistical test}; and (iii) a \textit{significance level}. 

\phead{Reference set}
Let 
$\mathcal{C}' = \{c'_1, c'_2, \dots, c'_{|\mathcal{C}'|}\}$ be a set 
of random concept mentions, called the \textit{reference set}, with $|\mathcal{C}'| = |\mathcal{C}|$. For each axis $j$, the concept projection component (\secref{sec:projection}) produces projections $\pi_j(c)$ for $c \in \mathcal{C}$ and $\pi_j(c')$ for $c' \in \mathcal{C}'$ (z-scored jointly over $\mathcal{C} \cup \mathcal{C}'$).

\phead{Statistical test}
We use the two-sample Kolmogorov--Smirnov 
(KS) test \citep{kolmogorov1933, smirnov1933}, a non-parametric test that compares two empirical distributions without assumptions about their underlying 
distributions. 
For each semantic axis $j$, the test yields a KS statistic $y_j \in [0,1]$ (the maximum distance between the two distributions) and a $p$-value $p_j$. The null hypothesis ($H_0$) is that $\pi_j(c)$ and $\pi_j(c')$ follow the same distribution; the alternative ($H_1$) is that they follow different distributions.


%
\phead{Significance level} We reject $H_0$ and classify 
axis $j$ as stereotypical whenever $p_j < \alpha = 0.05$. The KS statistic $y_j$ also serves as the stereotypicality score for each semantic axis.

\section{Case Study: Social Group Stereotypes}
\label{sec:comparison}

Having introduced \ours, a framework that can discover semantic axes in LLMs and quantify their stereotypicality with respect to relevant concepts (H1-H3), we now apply it to the case study of social group stereotypes to test H4. Two complementary comparisons make this test concrete: \textit{(a) Do \ours's projections of social
groups onto the geometric axes discovered by \ours align with human ratings on the corresponding semantic axes?} (\secref{sec:rq2_positions}) and \textit{(b) Do the semantic axes which \ours identifies as stereotypical align with those identified by humans?} (\secref{sec:rq2_axes}). 

For our experiments, we use the instruction-tuned variants of the 
open-weight decoder-only LLMs \llama~\citep{llama3} and 
\mistral~\citep{mistral7b}, which give us both instruction-following behavior and access to internal 
representations. For head aggregation (Eq.~\ref{eq:agg_projection}), we 
use $k = 128$ unless stated otherwise, and fix the random seed to 42 across all experiments.

\subsection{Position Prediction}
\label{sec:rq2_positions}

\phead{Data: social psychology surveys} 
We require ground truth for social group ratings along SCM dimensions. We use ground-truth labels that are well established across the social psychology literature, rather than ratings obtained from a small annotator pool in a single study. We
use the compilation of \citet{fraser-etal-2021-understanding}, who aggregated Warmth and Competence labels for 25 social group mentions. To construct the semantic
axes, we draw on the stereotype dictionary of
\citet{nicolas_2021}, which organizes antonymous adjective pairs under
each facet of Competence and Warmth (e.g., \vs{unfriendly.a.02}{friendly.a.01} under
Sociability), yielding 50 Warmth and 42 Competence semantic axes. Each social group is thus
assigned a binary label ($\pm 1$) on each semantic axis. Full
lists of semantic axes and social groups are in
Appendix~\ref{app:casestudy_datasets}. 

\phead{Setup}
For social group $g$ and axis $j$, \ours produces a scalar projection
$\pi_j(g)$ (Eq.~\ref{eq:projection}), whose sign gives the predicted
label: positive projections map to $+1$, and negative projections
to $-1$. We refer to this as \textit{position prediction}, and use
\textit{accuracy} -- the proportion of predictions
that agree with the ground-truth labels -- as the evaluation metric, averaged over axes within the Warmth and Competence categories.

\begin{figure}[!t]
\centering
\begin{minipage}{0.49\columnwidth}
\centering
\resizebox{\linewidth}{!}{%
\begin{tikzpicture}[font=\scriptsize]
\node[font=\small\bfseries] at (1, 2.3) {Warmth};
\foreach \i/\lab in {0/Human, 1/Llama, 2/Mistral}{
  \node at (\i*1.0, 1.85) {\lab};
}
\foreach \i/\lab in {0/Human, 1/Llama, 2/Mistral}{
  \node at (-0.8, 1.4-\i*0.55) {\lab};
}
\foreach \i/\j/\v/\pct in {
  0/0/1.00/100, 0/1/0.57/57, 0/2/0.55/55,
  1/0/0.57/57, 1/1/1.00/100, 1/2/0.73/73,
  2/0/0.55/55, 2/1/0.73/73, 2/2/1.00/100}{
  \node[
    fill=stereoOrange!\pct,
    draw=black!30,
    minimum width=0.85cm,
    minimum height=0.5cm
  ] at (\j*1.0, 1.4-\i*0.55) {\v};
}
\end{tikzpicture}%
}
\end{minipage}%
\begin{minipage}{0.49\columnwidth}
\centering
\resizebox{\linewidth}{!}{%
\begin{tikzpicture}[font=\scriptsize]
\node[font=\small\bfseries] at (1, 2.3) {Competence};
\foreach \i/\lab in {0/Human, 1/Llama, 2/Mistral}{
  \node at (\i*1.0, 1.85) {\lab};
}
\foreach \i/\lab in {0/Human, 1/Llama, 2/Mistral}{
  \node at (-0.8, 1.4-\i*0.55) {\lab};
}
\foreach \i/\j/\v/\pct in {
  0/0/1.00/100, 0/1/0.63/63, 0/2/0.62/62,
  1/0/0.63/63, 1/1/1.00/100, 1/2/0.75/75,
  2/0/0.62/62, 2/1/0.75/75, 2/2/1.00/100}{
  \node[
    fill=stereoOrange!\pct,
    draw=black!30,
    minimum width=0.85cm,
    minimum height=0.5cm
  ] at (\j*1.0, 1.4-\i*0.55) {\v};
}
\end{tikzpicture}%
}
\end{minipage}
\caption{Pairwise agreement (position prediction accuracy \S\ref{sec:rq2_positions}) between Human,
\llama, and \mistral. The two LLMs agree with each other
more than with human ratings of social groups on stereotypical semantic axes.}
\label{fig:rq2_3x3_agreement}
\end{figure}

\phead{Finding: LLMs align with humans better on Competence than on Warmth, and align more with each other than with humans} 
Fig.~\ref{fig:rq2_3x3_agreement} reports pairwise agreement between
Human, \llama, and \mistral. Both models align with
humans more strongly on Competence (Human--Llama: $0.63$; Human--Mistral: $0.62$)
than on Warmth (Human--Llama: $0.57$; Human--Mistral: $0.55$),
suggesting that Competence-related stereotypes are more strongly linearly encoded in
the LLM activation space than Warmth-related stereotypes. Interestingly, the Llama--Mistral agreement ($0.75$
Competence; $0.73$ Warmth) exceeds the human--LLM agreement
in both stereotype categories.

\begin{figure}[!t]
  \centering
  \includegraphics[width=\linewidth]{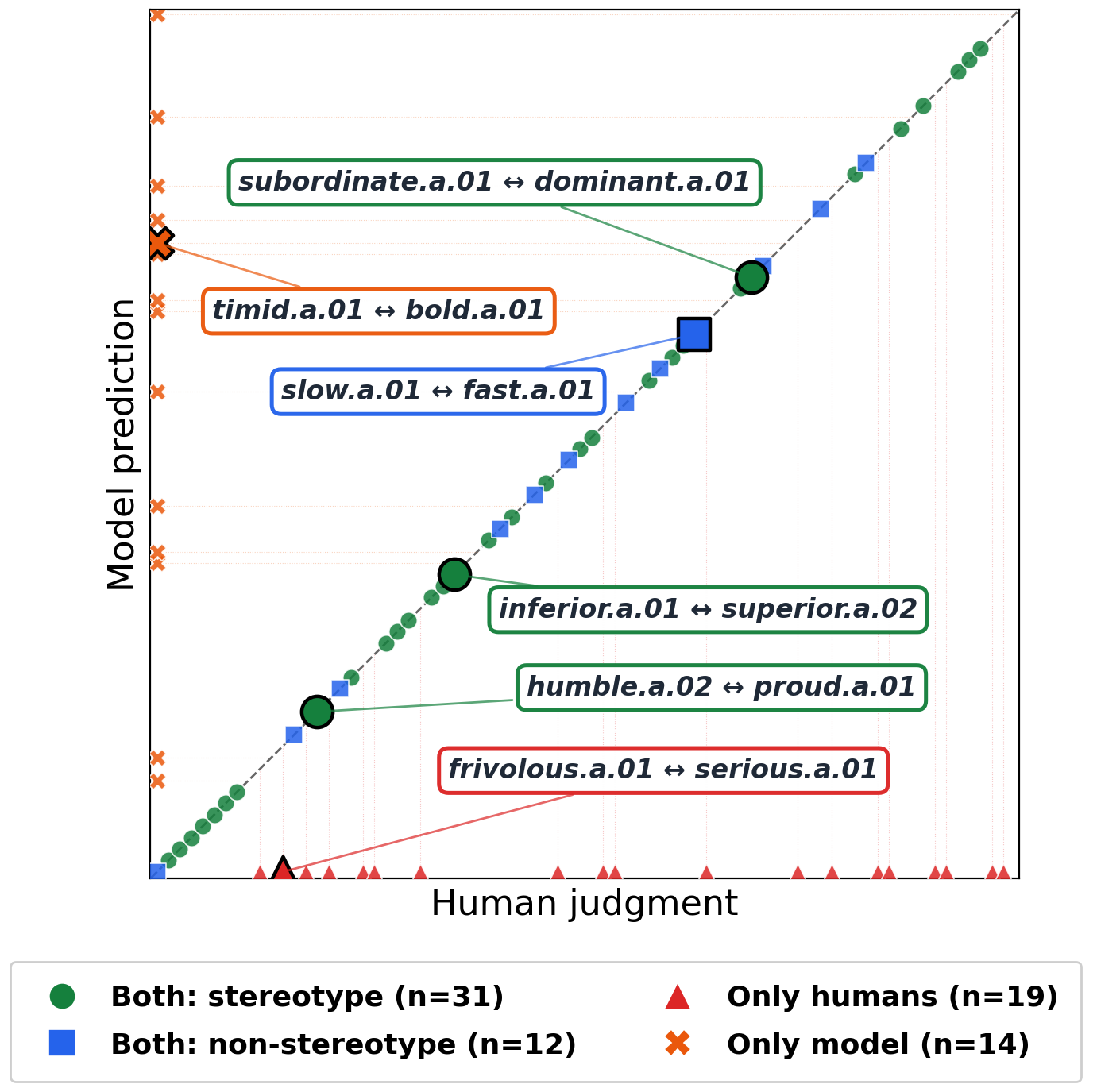}
  \caption{\ours instantiated with \llama compared against human annotations on a stratified sample of semantic axes, judging whether each axis is stereotypical. Each point is one of the 76 semantic axes, positioned by human judgment (x-axis) and \ours prediction (y-axis); diagonal points indicate agreement (43 axes: 31 stereotypical, 12 non-stereotypical). Off-diagonal points show disagreements (19 humans-only, 14 \ours-only).}
  \label{fig:diagonal}
\end{figure}

\subsection{Stereotypicality identification}
\label{sec:rq2_axes}

\begin{figure*}[!t]
  \centering
  \begin{subfigure}[t]{0.49\linewidth}
    \centering
    \includegraphics[width=\linewidth]{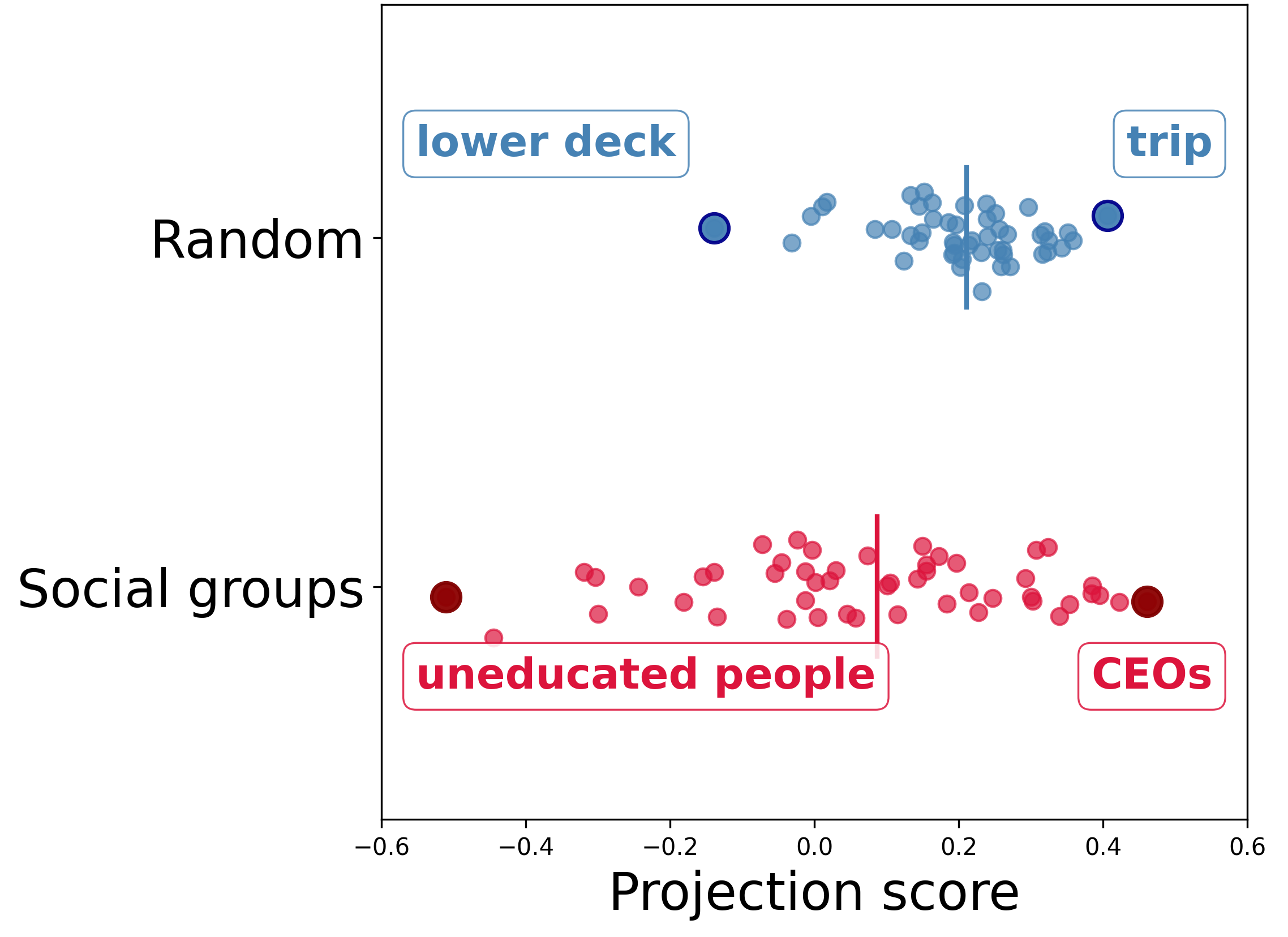}
    \caption{Stereotypical axis \vs{inferior.a.01}{superior.a.02}: social groups spread to both poles; random phrases cluster near center.}
    \label{fig:rq2_proj_stereo}
  \end{subfigure}
  \hfill
  \begin{subfigure}[t]{0.49\linewidth}
    \centering
    \includegraphics[width=\linewidth]{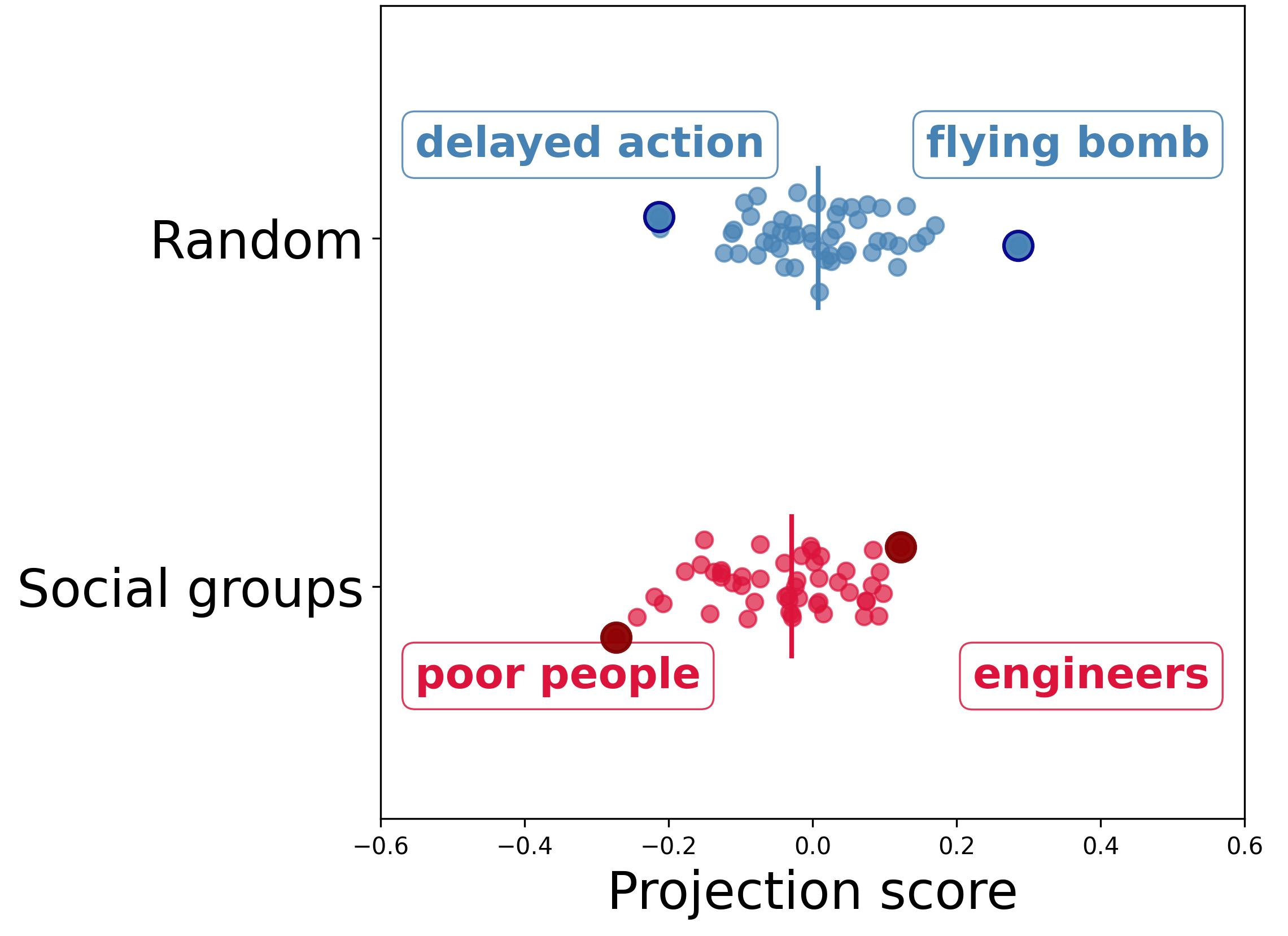}
    \caption{Non-stereotypical axis \vs{slow.a.01}{fast.a.01}: social groups and random phrases overlap with near-identical medians.}
    \label{fig:rq2_proj_nonstereo}
  \end{subfigure}
  \caption{Projection distributions for \ours with \llama. Each point is a social group mention (red) or random phrase (blue) projected onto the geometric axis (Eq.~\ref{eq:agg_projection}); vertical lines mark group medians.}
  \label{fig:rq2_projections}
\end{figure*}


\phead{Data: human stereotypicality annotations} We construct a stratified random sample of $100$
semantic axes from WordNet antonym synsets, covering three strata: \emph{Philosophy and
Psychology} ($34$ axes), \emph{Unknown} ($33$, axes without a
BabelDomains~\cite{camacho_collados} label), and \emph{Technical domains} ($33$). Human annotators provide judgments  for each axis  (see annotation details in
Appendix~\ref{app:casestudy_datasets}). After filtering axes marked as
\emph{Not applicable to humans} by a majority of annotators, 76 semantic axes remain. 
We use $50$ social group mentions for concept set $\concepts$
(Table~\ref{tab:social-groups-dataset2}) and $50$ frequency-matched random phrases for $\randomconcepts$ (Table~\ref{tab:random-nouns}). We control for frequency because it can affect stereotypical associations \cite{100_years_of_stereotypes}.  We compare \ours against aggregated human annotations to determine whether a semantic axis is stereotypical.

\phead{Setup} For each semantic axis $j$, \ours produces a stereotypicality score $y_j \in [0, 1]$ from the Kolmogorov--Smirnov test described in \secref{sec:hypothesis}, implemented using the \href{https://docs.scipy.org/doc/scipy/reference/generated/scipy.stats.ks_2samp.html}{SciPy} library, and classifies axis $j$ as stereotypical when $p_j < 0.05$. To assess whether our sample size ($n_{\mathcal{C}} = n_{\mathcal{C}'} = 50$) provides sufficient statistical power for detecting differences between the two distributions, we additionally conduct a power analysis reported in Appendix~\ref{app:power_analysis}. 


\phead{Finding 1: \ours overlaps with human judgments in identifying stereotypical semantic axes and additionally identifies stereotypical semantic axes not investigated in prior work}
Fig.~\ref{fig:diagonal} compares \ours instantiated with \llama
against human annotations across the 76 semantic axes. Points on the diagonal indicate agreement; off-diagonal points mark disagreements where an axis is flagged as stereotypical only by humans or only by \ours. For \llama, \ours agrees with human judgments on
43 of 76 axes (57\%); for \mistral, agreement is 45 of 76 (59\%).
Table~\ref{tab:novel} lists stereotypical semantic axes that \ours identifies and human annotators independently confirm, but which fall outside the semantic axes operationalized in prior research on stereotypes (\secref{sec:psy-model}; \citealp{nicolas_2021}). 

\phead{Finding 2: Stereotypical semantic axes exhibit distinct projection distributions}
For a stereotypical semantic axis such as \vs{inferior.a.01}{superior.a.02} (Fig.~\ref{fig:rq2_proj_stereo}), social group mentions spread across a wider range of projection scores than random phrases, reaching toward both poles: ``uneducated people'' falls on the negative pole and ``CEOs'' on the positive pole, while random phrases stay close to the middle. For a non-stereotypical semantic axis such as \vs{slow.a.01}{fast.a.01} (Fig.~\ref{fig:rq2_proj_nonstereo}), the social group mentions and random phrases cover the same range of projection scores with close medians -- the LLM does not differentiate social groups from random phrases along that axis.

\begin{table}[t]
\centering
\footnotesize
\renewcommand{\arraystretch}{1.2}
\setlength{\tabcolsep}{6pt}
\begin{tabular}{@{}p{4.0cm} cc@{}}
\toprule
\textbf{Semantic axis (Neg./Pos.)} & \textbf{Llama} & \textbf{Mistral} \\
\midrule
unreasonable.a.01/reasonable.a.01      & 0.62 & 0.44 \\
cowardly.a.01/brave.a.01               & 0.56 & 0.32 \\
narrow-minded.a.02/broad-minded.a.02   & 0.34 & 0.60 \\
retarded.a.01/precocious.a.01          & 0.32 & 0.40 \\
humble.a.02/proud.a.01                 & 0.28 & 0.16$^\dagger$ \\
\bottomrule
\end{tabular}
\caption{Five stereotypical semantic axes discovered by \ours beyond the stereotype dictionary \cite{nicolas_2021}. Values are KS statistics for each LLM. Llama refers to \llama and Mistral refers to \mistral. All entries are significant at $p < 0.05$ except the one marked with~$\dagger$, which is not considered stereotypical.}
\label{tab:novel}
\end{table}

\section{Discussion}

The case study primarily tests H4 (human alignment), but its results also provide empirical evidence consistent with the framework hypotheses H1 (geometric encoding), H2 (rating-as-projection) and H3 (distributional shift). Position prediction accuracy (\secref{sec:rq2_positions}) is consistent with H1 and H2: \ours's projections recover human SCM labels above chance, more strongly on Competence (0.62-0.63) than on Warmth (0.55-0.57). Stereotypicality identification (\secref{sec:rq2_axes}) is consistent with H3: the KS test flags 57-59\% of the axes that human annotators also flag, and stereotypical axes exhibit clearly distinct projection distributions (Fig.~\ref{fig:rq2_proj_stereo}) while non-stereotypical axes overlap (Fig.~\ref{fig:rq2_proj_nonstereo}). \ours additionally identifies stereotypical axes outside the dictionary used in prior work (Table~\ref{tab:novel}) -- including \vs{cowardly}{brave}, \vs{narrow-minded}{broad-minded} and \vs{humble}{proud}.
H4 (human alignment) is only partially supported. \ours agrees with humans on 55-63\% of axis positions and on 57-59\% of stereotypicality classifications. More striking, \llama and \mistral agree with each other (0.73-0.75 on positions) more than either does with humans - the two LLMs share stereotype content that diverges from what is documented in social psychology. Cross-model consistency therefore cannot be taken as evidence that LLM stereotypes match human ones.

These findings have direct consequences for stereotype mitigation in LLMs: approaches that target only predefined semantic axes 
\citep{omrani, li-etal-2025-fairsteer} leave axes such as
\vs{retarded}{precocious} untouched.
Because \ours identifies the specific attention heads that linearly
encode each stereotypical axis, the recovered geometric axes are directly
usable for steering LLM outputs \citep{li_intervention} -- without requiring additional prompting or fine-tuning. 

\section{Conclusion}

We introduced \ours, a framework that adapts the semantic
differential method to the study of stereotypes in LLM internal representations. By recovering semantic axes as geometric axes in
activation space and applying a statistical test over concept
projections, \ours enables systematic exploration of candidate stereotype
axes without relying on a predefined axis dictionary. Applied to
social group stereotypes, \ours reveals that the stereotypes LLMs encode
only partially align with those documented in social psychology and discovers stereotypical semantic axes outside the dictionary used in prior
 work -- findings with direct consequences
for how LLM stereotypes are scoped for auditing and mitigation.


\newpage
\section*{Limitations}
For the case study, we chose social group stereotypes because this
domain offers widely accepted ground truth labels, compiled across
decades of psychological surveys (\S\ref{app:psychology_surveys})
rather than from any single study. \ours
itself is general and can be applied to other concept families such
as brands~\citep{brands_stereotypes}, food
items~\citep{kozlowski2019geometry}, or individuals, but empirical experiments in those domains require 
comparable ground truth, which remains an open direction for future work.

\ours requires that the chosen semantic axes be meaningful for the concept 
under study. For social groups, this means restricting attention to axes
that can describe human attributes; axes drawn from domains
such as geometry (e.g., \vs{asymmetric}{symmetric}) or chemistry 
(e.g., \vs{acidic}{alkaline})
are not applicable and produce uninformative projections. Our stratified sampling procedure (\S\ref{app:stratified})
partially controls for this, and the rest is controlled based on human 
annotation (\S\ref{app:annotation_aggregation}), but in general 
the selection of applicable semantic axes remains the responsibility of the analyst.

In this paper, we study LLM stereotypes, which may differ from 
human stereotypes \citep{NicolasCaliskan2025ChatbotStereotypes}. 
Importantly, no ground truth exists for what stereotypes a given 
LLM ``should'' encode, making it difficult to directly evaluate 
whether \ours captures LLM stereotypes more faithfully than 
alternative methods (\S\ref{app:ablation}). As a proxy, we compare 
\ours outputs against human stereotypes drawn from social psychology 
surveys and our annotation effort.  These human labels themselves reflect the cultural
assumptions of their respective annotator pools, which are
predominantly Western and educated. Consequently, divergences between
\ours and human labels may reflect genuine differences between LLM
and human stereotypes rather than failures of the framework.

Our experiments cover two open-weight 7--8B instruction-tuned models 
from different developers: \llama, developed by Meta, 
and \mistral, developed by Mistral AI. Despite their 
similar size, the two models differ in architecture, training data, and 
alignment procedure, making them a representative pair for studying 
whether findings generalize across models. How the alignment between human and LLM stereotypes shifts across 
substantially larger models or models trained under different objectives 
remains an open question and constitutes a promising direction for 
future work.

Our current study is limited to English, as the sentence templates are in English and the opposites used to construct semantic axes (\S\ref{sec:constructing-axis}) are derived from WordNet. However, extending \ours beyond English should be feasible, as the sentence templates can be adapted to other languages and WordNet can be replaced with multilingual resources such as ConceptNet~\citep{conceptnet} or BabelNet~\citep{babelnet}.

\section*{Ethical Considerations}

This paper studies stereotypes encoded in LLMs with the goal of making such associations more transparent. Because our analysis necessarily surfaces stereotypical and potentially offensive content, we present these examples strictly for analysis and not as endorsement. Our case study draws on a non-exhaustive set of social groups from prior work, together with stereotypicality judgments collected from volunteer annotators with diverse linguistic and disciplinary backgrounds. Annotators were informed of the sensitive nature of the task and consented to participation in the study. No personally identifying information or sensitive personal data was collected, no vulnerable populations were involved, and formal ethics review board approval was therefore not required.

Because \ours recovers geometric axes associated with stereotypical semantic axes, the recovered geometric axes could potentially be used for activation steering that amplifies stereotypical behavior in LLM generations. This possibility motivates careful consideration of the risks associated with applying \ours.




\bibliography{main}

\clearpage
\appendix

\begin{table*}[h!]
\centering
\small
\setlength{\tabcolsep}{4pt}
\renewcommand{\arraystretch}{1.1}
\scalebox{0.78}{%
\begin{tabular}{r p{7.5cm} r p{7.5cm}}
\toprule
\multicolumn{4}{l}{\textit{Simple templates (30 templates)}} \\
\midrule
1  & \{ADJ\} is: \{DEF\}.
 & 16 & A person described as \{ADJ\} is \{DEF\}. \\
2  & \{ADJ\} means: \{DEF\}.
 & 17 & \{ADJ\} can be defined as \{DEF\}. \\
3  & Definition of \{ADJ\}: \{DEF\}.
 & 18 & The concept of \{ADJ\}: \{DEF\}. \\
4  & \{ADJ\}: \{DEF\}.
 & 19 & \{ADJ\} in one word: \{DEF\}. \\
5  & The word \{ADJ\} means \{DEF\}.
 & 20 & Simply put, \{ADJ\} means \{DEF\}. \\
6  & The adjective \{ADJ\} is defined as: \{DEF\}.
 & 21 & \{ADJ\} = \{DEF\}. \\
7  & \{ADJ\} refers to: \{DEF\}.
 & 22 & What does \{ADJ\} mean? \{DEF\}. \\
8  & Meaning of \{ADJ\}: \{DEF\}.
 & 23 & \{ADJ\}, which means \{DEF\}. \\
9  & \{ADJ\} --- \{DEF\}.
 & 24 & The meaning of \{ADJ\} is \{DEF\}. \\
10 & The term \{ADJ\} describes: \{DEF\}.
 & 25 & If someone is \{ADJ\}, they are \{DEF\}. \\
11 & \{ADJ\}, i.e., \{DEF\}.
 & 26 & \{ADJ\}: essentially \{DEF\}. \\
12 & \{ADJ\} (meaning \{DEF\}).
 & 27 & In short, \{ADJ\} means \{DEF\}. \\
13 & Something \{ADJ\} is \{DEF\}.
 & 28 & \{ADJ\} --- in other words, \{DEF\}. \\
14 & To be \{ADJ\} is to be \{DEF\}.
 & 29 & To call something \{ADJ\} is to say it is \{DEF\}. \\
15 & \{ADJ\}: characterized by \{DEF\}.
 & 30 & \{ADJ\}, that is, \{DEF\}. \\
\midrule
\multicolumn{4}{l}{\textit{Listing templates (30 templates)}} \\
\midrule
1  & List adjectives similar to \{ADJ\}, in the sense of \{DEF\}.
 & 16 & Provide adjectives one would use to recognize something as \{ADJ\}, meaning \{DEF\}. \\
2  & Name synonyms of \{ADJ\}, meaning \{DEF\}.
 & 17 & List the typical features of something that is \{ADJ\} i.e., \{DEF\}. \\
3  & Provide words similar to \{ADJ\} i.e., \{DEF\}.
 & 18 & Name adjectives that reflect \{ADJ\}, in the sense of \{DEF\}. \\
4  & Name adjectives that share the meaning of \{ADJ\}, in the sense of \{DEF\}.
 & 19 & List adjectives that reflect observable qualities of \{ADJ\}, meaning \{DEF\}. \\
5  & List adjectives that capture the condition of being \{ADJ\}, meaning \{DEF\}.
 & 20 & Provide adjectives that make something feel distinctly \{ADJ\} i.e., \{DEF\}. \\
6  & Provide similar adjectives to \{ADJ\} i.e., \{DEF\}.
 & 21 & Name defining properties commonly linked to \{ADJ\}, in the sense of \{DEF\}. \\
7  & Name adjectives that reflect the state of being \{ADJ\}, in the sense of \{DEF\}.
 & 22 & List adjectives that describe how something earns the description \{ADJ\}, meaning \{DEF\}. \\
8  & List closely related adjectives to \{ADJ\}, meaning \{DEF\}.
 & 23 & Provide the qualities implied when something is said to be \{ADJ\} i.e., (\{DEF\}). \\
9  & Provide similar words to \{ADJ\}, in the sense of \{DEF\}.
 & 24 & Name adjectives that convey the nature of \{ADJ\}, in the sense of \{DEF\}. \\
10 & Name adjectives that describe manifestations of \{ADJ\}, meaning \{DEF\}.
 & 25 & List the shared traits among instances described as \{ADJ\}, meaning \{DEF\}. \\
11 & List \{ADJ\} characteristics i.e., \{DEF\}.
 & 26 & Provide the defining signals that indicate \{ADJ\} i.e., \{DEF\}. \\
12 & Provide three synonyms of the adjective \{ADJ\}, in the sense of \{DEF\}.
 & 27 & Name the qualities that justify calling something \{ADJ\}, in the sense of \{DEF\}. \\
13 & Name adjectives that describe how \{ADJ\} presents itself, meaning \{DEF\}.
 & 28 & List the qualities that make an example clearly \{ADJ\}, meaning \{DEF\}. \\
14 & List adjectives for situations that would be considered \{ADJ\} i.e., \{DEF\}.
 & 29 & Provide adjectives that create the general impression of \{ADJ\} i.e., \{DEF\}. \\
15 & Name behaviors or traits that exemplify being \{ADJ\}, in the sense of \{DEF\}.
 & 30 & Name adjectives that describe the defining aspects of \{ADJ\}, in the sense of \{DEF\}. \\
\bottomrule
\end{tabular}
}
\caption{All templates used to construct the probing dataset, divided
into two categories: \textit{Simple} templates, which elicit
a small number of synonyms or related words, and \textit{Listing}
templates, which explicitly request a list of adjectives,
characteristics, or qualities and thus expose the model to multiple
synonyms of the target adjective in a single context. \{ADJ\} is
replaced by the adjective synset and \{DEF\} by its WordNet definition,
which disambiguates the sense. The ablation in
Appendix~\ref{app:ablation} (Table~\ref{tab:ablation})
shows that \textit{Listing} templates outperform \textit{Simple}
templates across both models and both dimensions; we therefore use
\textit{Listing} templates in the main experiments.}
\label{tab:templates}
\end{table*}

\section{Ablation study}
\label{app:ablation}

 Our framework needs four configuration choices in
\S\ref{sec:linear-probing} and \S\ref{sec:projection}: (i) the
template formulation used to instantiate the probing dataset
(Listing  vs.\ Simple ;
Table~\ref{tab:templates}), (ii) the number of templates
$n \in \{15, 30\}$ used to instantiate the probing dataset, (iii) the token position for
extracting activations (mean over all tokens vs.\ the last token), and
(iv) the ensemble size $k$ of attention heads aggregated
in Eq.~\ref{eq:agg_projection}. 
 
 This appendix reports a full ablation over the mentioned configuration choices, evaluated on the  dataset of social psychology surveys (\S~\ref{sec:rq2_positions}), against
the Warmth and Competence ground truth labels from the Stereotype
Content Model.
Table~\ref{tab:ablation} reports
position prediction accuracy for every combination, for both
\llama and \mistral. 

We note that in this paper, we study stereotypes encoded in LLMs, 
which may differ from human ones \citep{NicolasCaliskan2025ChatbotStereotypes}. 
No ground truth exists for what stereotypes a given LLM ``should'' encode, 
making it impossible to directly evaluate the correctness of \ours. 
For our ablation experiments, we therefore treat agreement with human 
ratings as a \textbf{signal} rather than a measure of \textbf{correctness}: 
it serves as a reference point for assessing how different configuration 
choices impact the output of \ours.

\paragraph{Template formulation.} Listing templates outperform
Simple templates across nearly every cell, with the gap
widening as $k$ grows. This pattern is particularly clear on
Competence: at $k = 512$, Listing reaches $.69$ on \llama and $.68$
on \mistral, against $.65$ and $.66$ for Simple. We attribute this
to the fact that Listing templates expose the LLM to multiple
synonyms of the target adjective in a single context, providing
the probing procedure with richer in-context evidence for the
pole's meaning.

\paragraph{Number of templates.} Doubling the number of templates
from $n = 15$ to $n = 30$ has a negligible effect: the average gain
across all cells is $+0.002$, with $96\%$ of cells differing by at
most $0.02$, indicating that $n = 15$ already provides sufficient template diversity. 

\begin{figure*}[h!]
\centering

\begin{minipage}{0.46\linewidth}\centering\textbf{\llama}\end{minipage}\hfill
\begin{minipage}{0.46\linewidth}\centering\textbf{\mistral}\end{minipage}

\vspace{0.15em}

\begin{minipage}{0.23\linewidth}\centering\small Last token\end{minipage}\hfill
\begin{minipage}{0.23\linewidth}\centering\small Mean over tokens\end{minipage}\hfill
\begin{minipage}{0.23\linewidth}\centering\small Last token\end{minipage}\hfill
\begin{minipage}{0.23\linewidth}\centering\small Mean over tokens\end{minipage}

\vspace{0.3em}

\begin{subfigure}{\linewidth}
\centering
\includegraphics[width=0.23\linewidth]{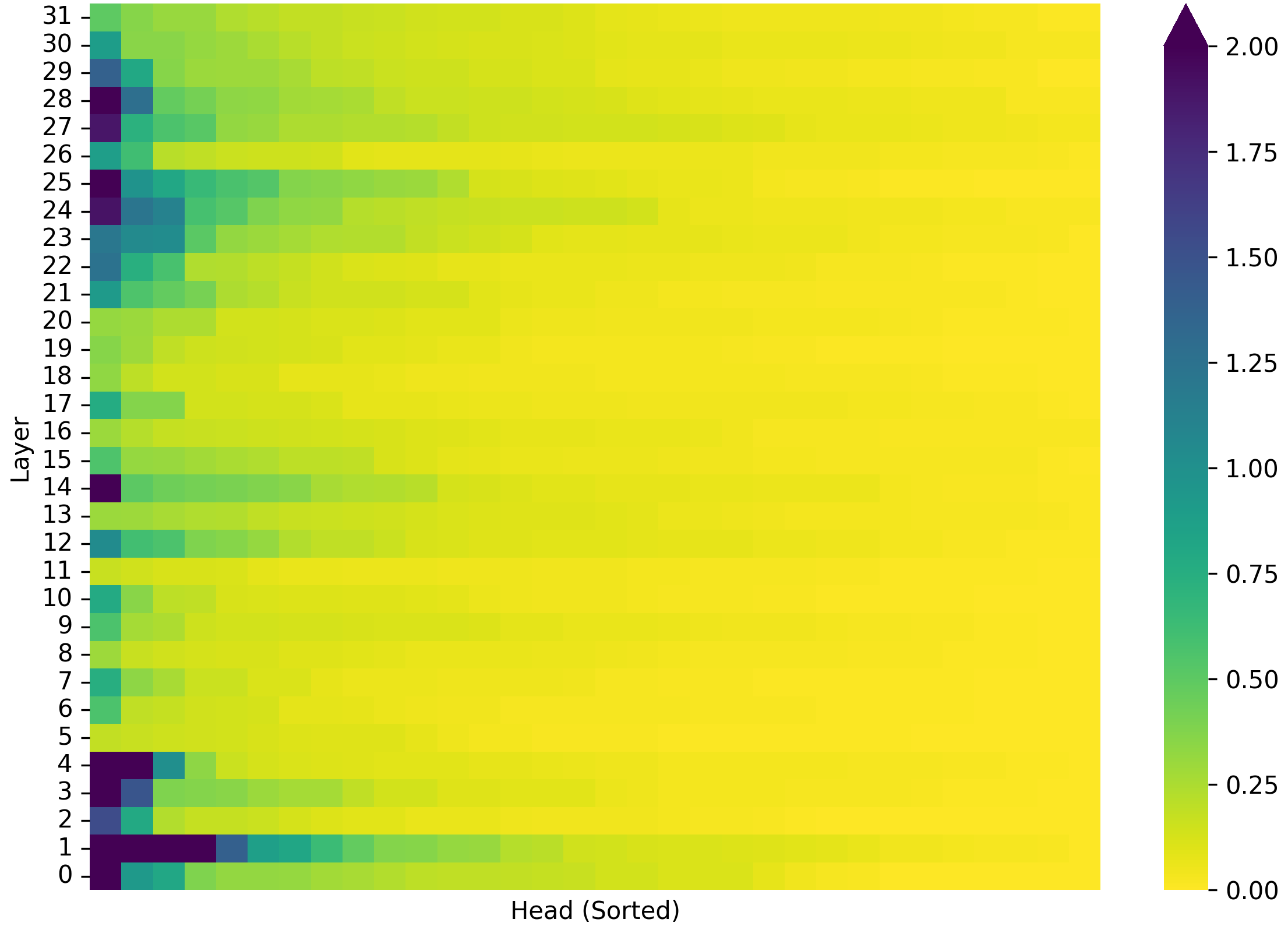}\hfill
\includegraphics[width=0.23\linewidth]{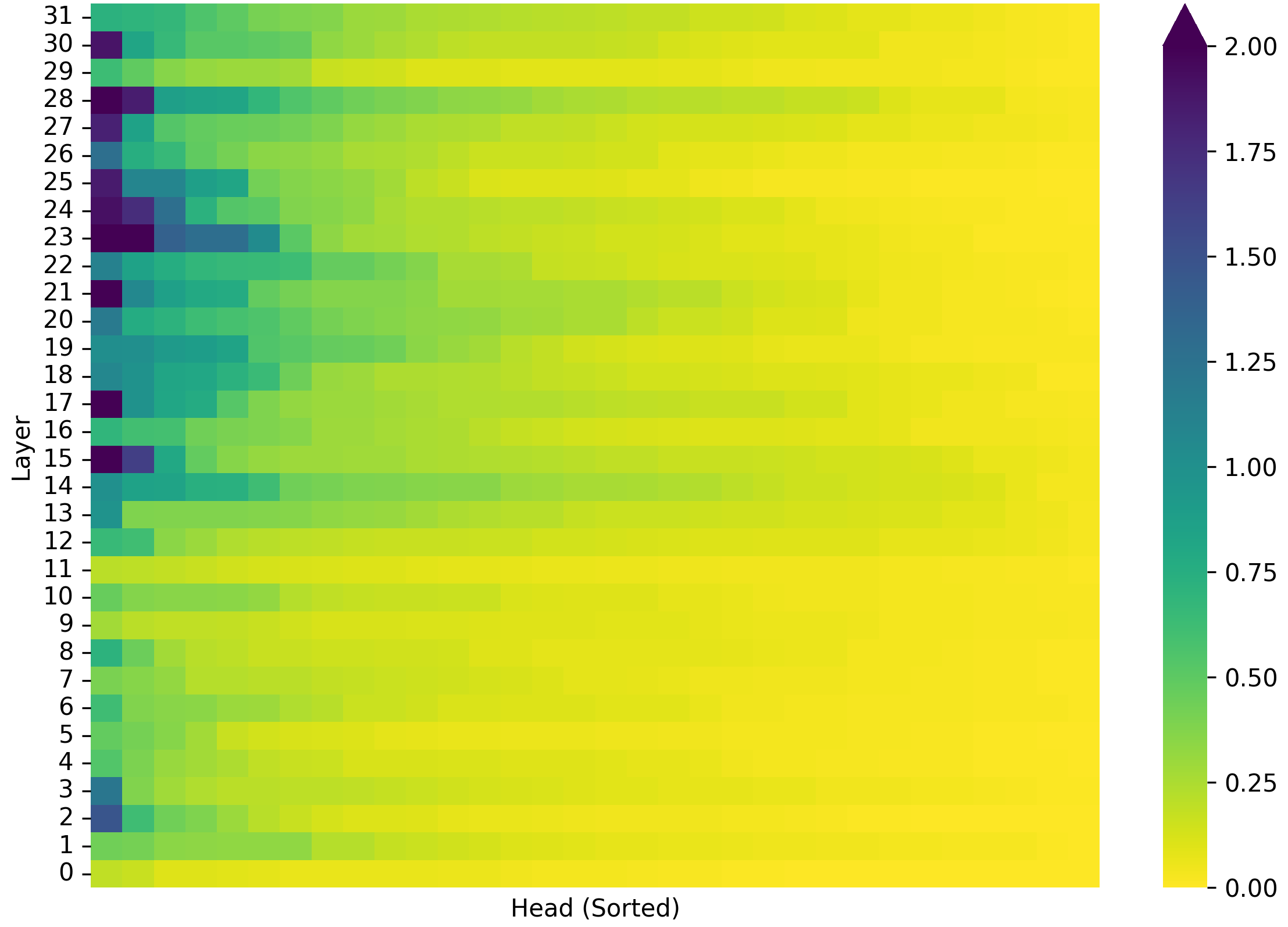}\hfill
\includegraphics[width=0.23\linewidth]{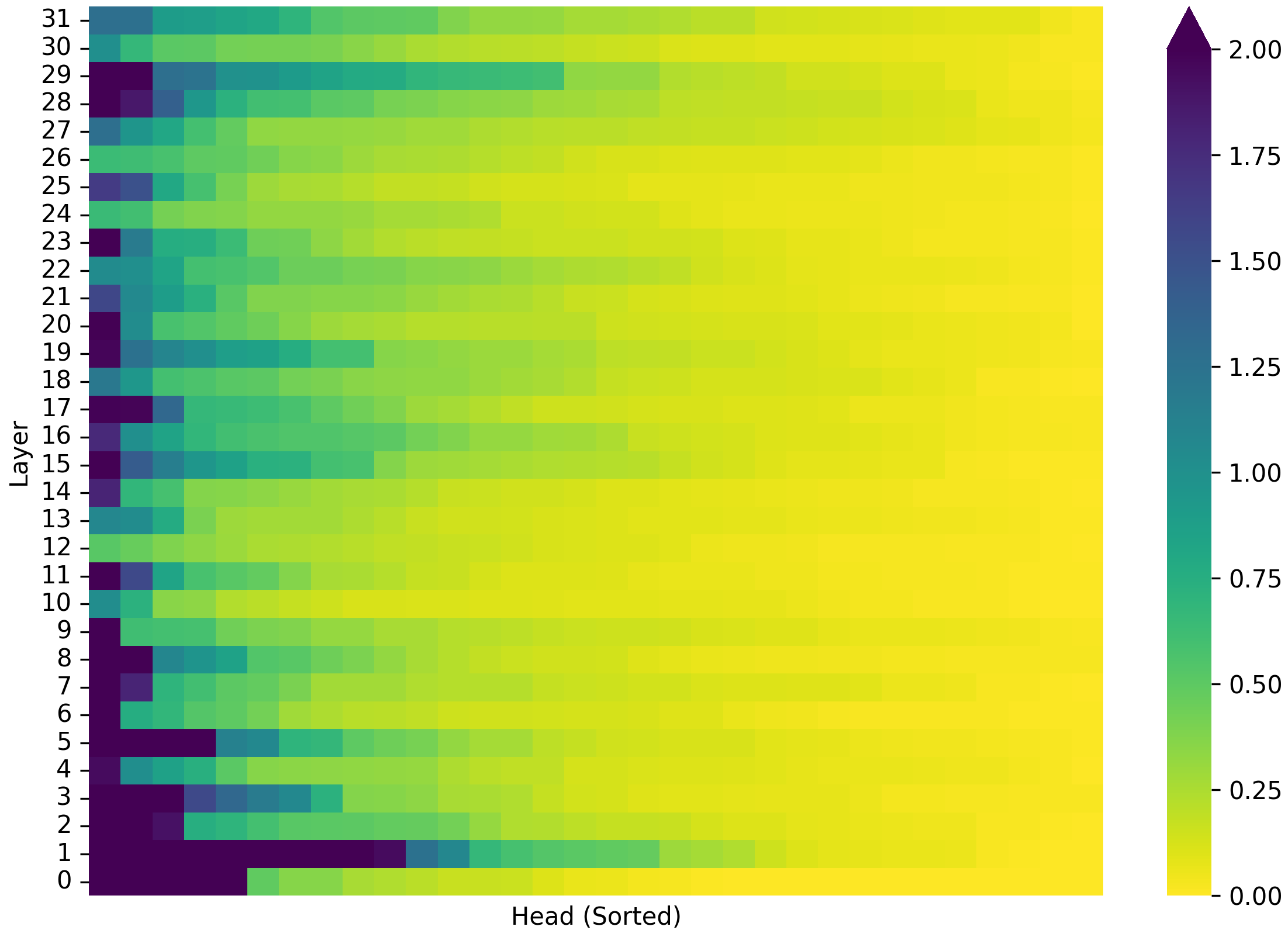}\hfill
\includegraphics[width=0.23\linewidth]{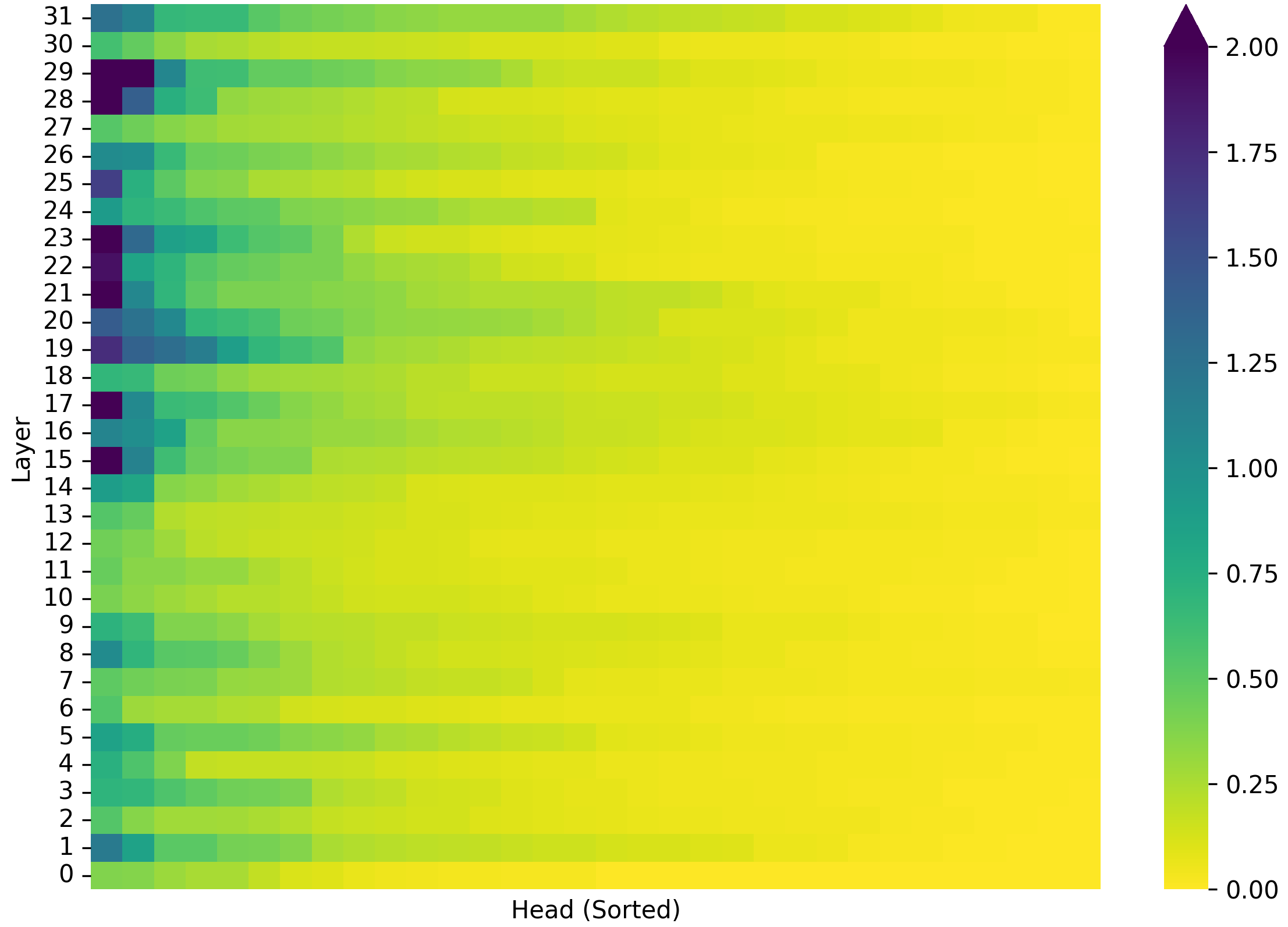}
\caption{\axis{subordinate.a.01}{dominant.a.01}}
\label{fig:tp_subordinate}
\end{subfigure}

\vspace{0.6em}

\begin{subfigure}{\linewidth}
\centering
\includegraphics[width=0.23\linewidth]{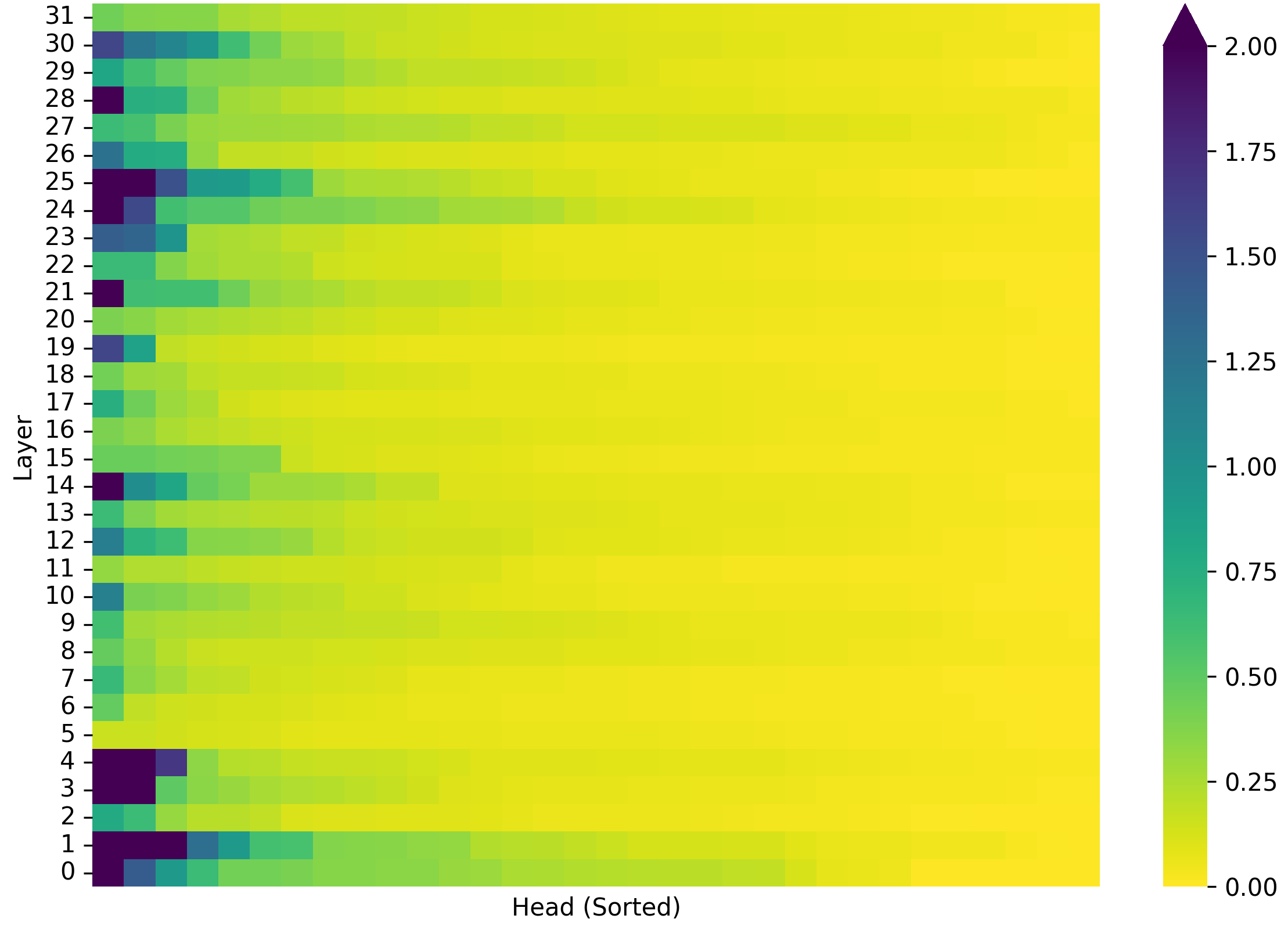}\hfill
\includegraphics[width=0.23\linewidth]{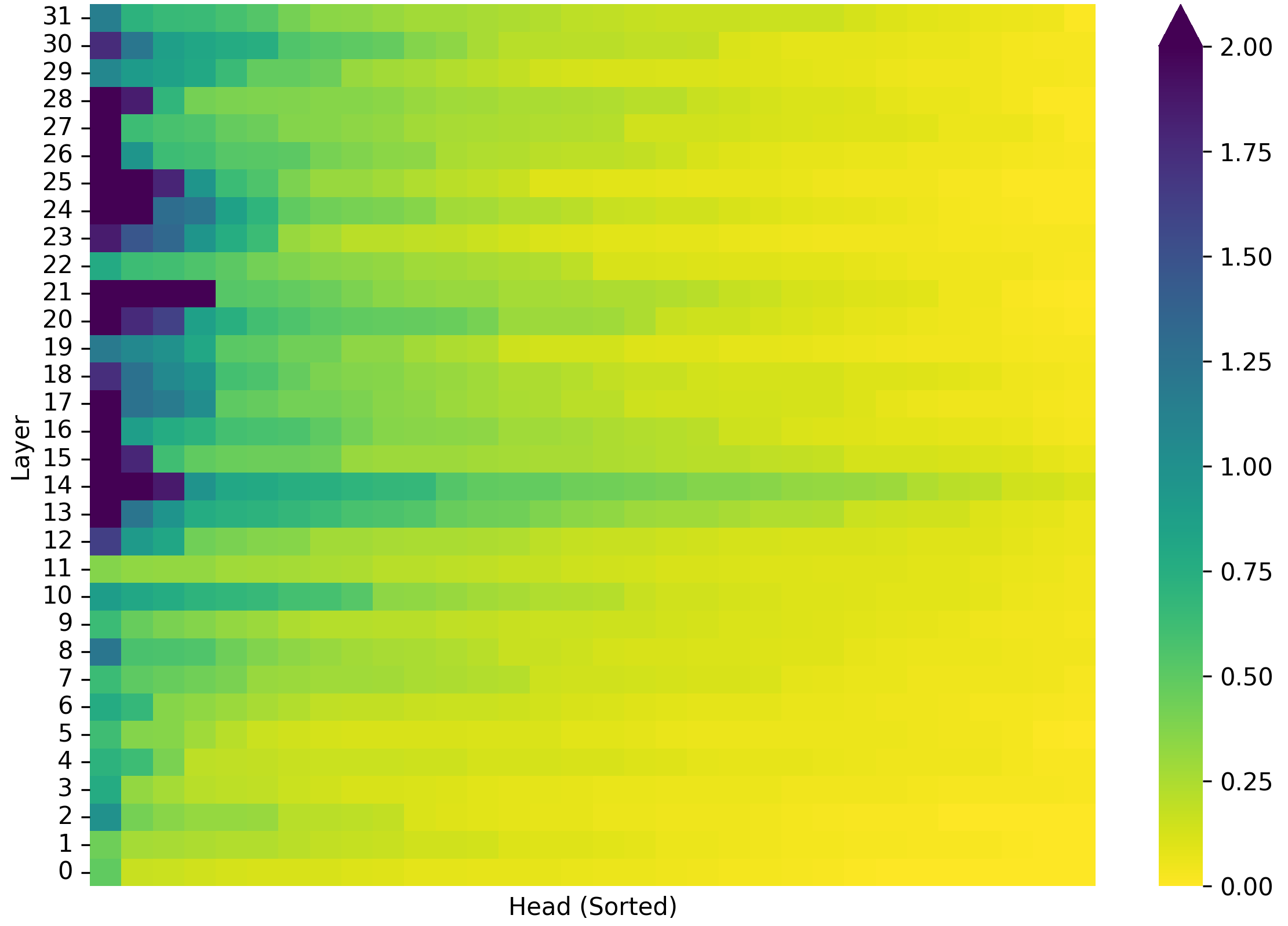}\hfill
\includegraphics[width=0.23\linewidth]{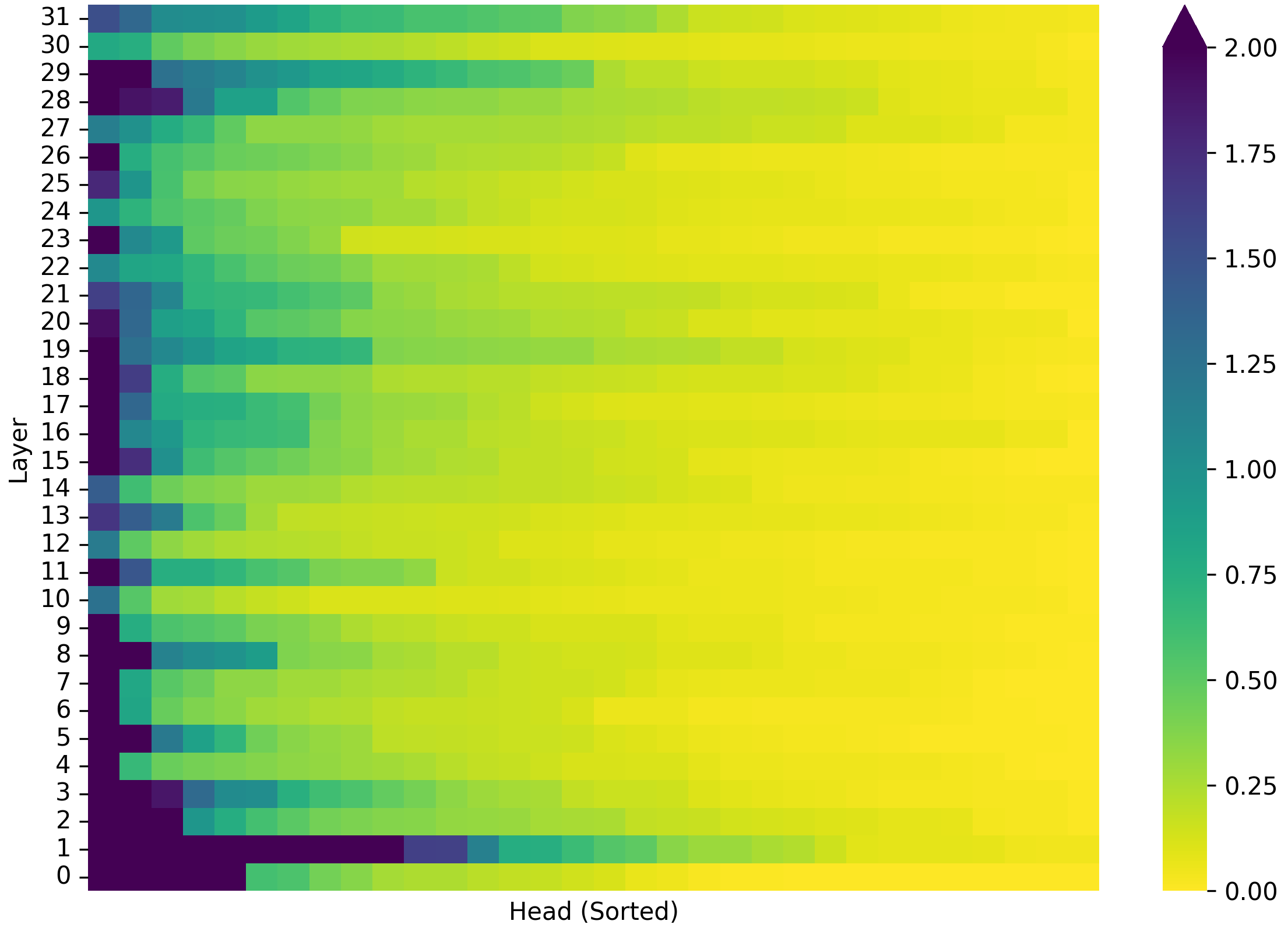}\hfill
\includegraphics[width=0.23\linewidth]{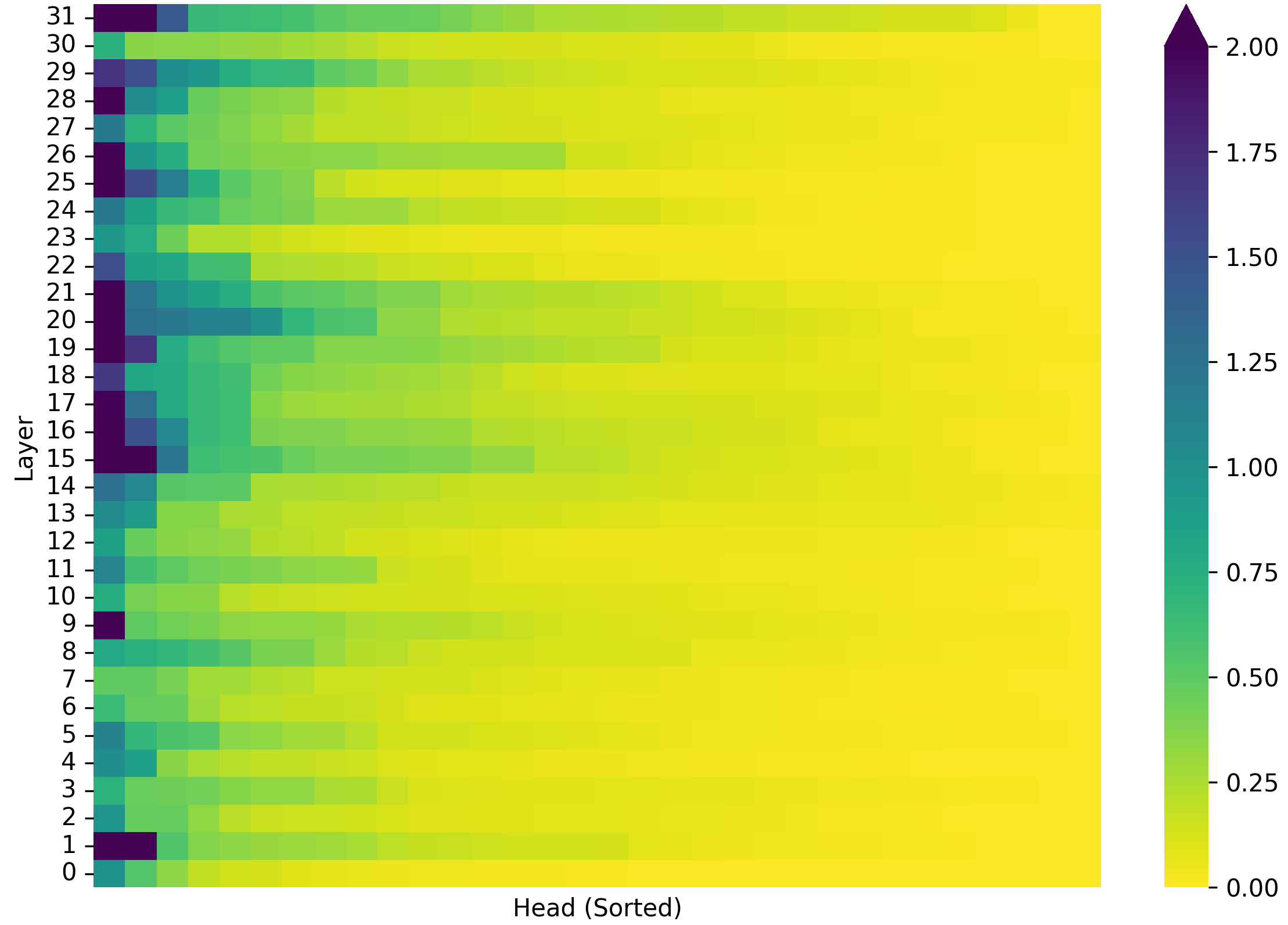}
\caption{\axis{timid.a.01}{bold.a.01}}
\label{fig:tp_timid}
\end{subfigure}





\vspace{0.6em}

\begin{subfigure}{\linewidth}
\centering
\includegraphics[width=0.23\linewidth]{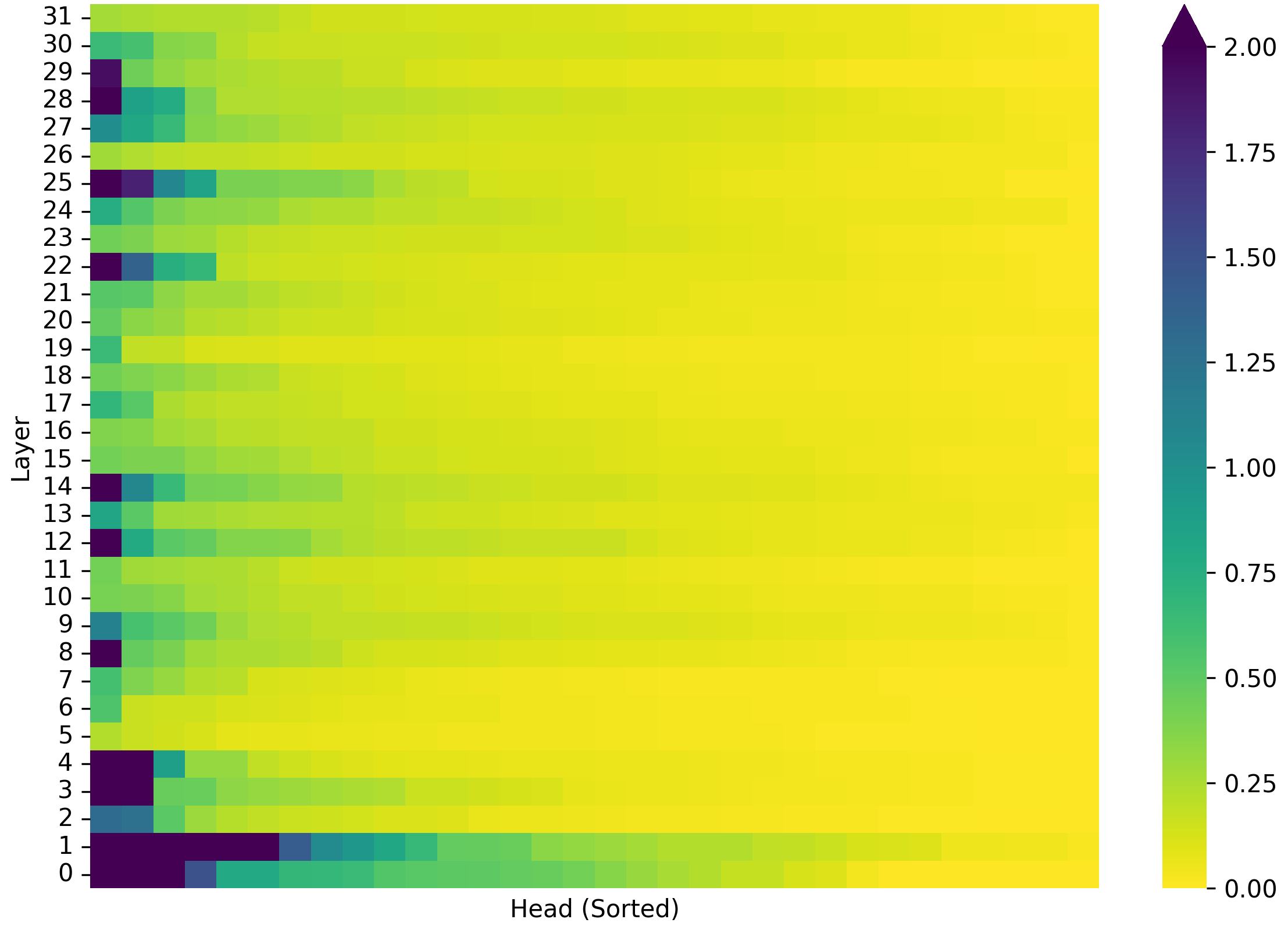}\hfill
\includegraphics[width=0.23\linewidth]{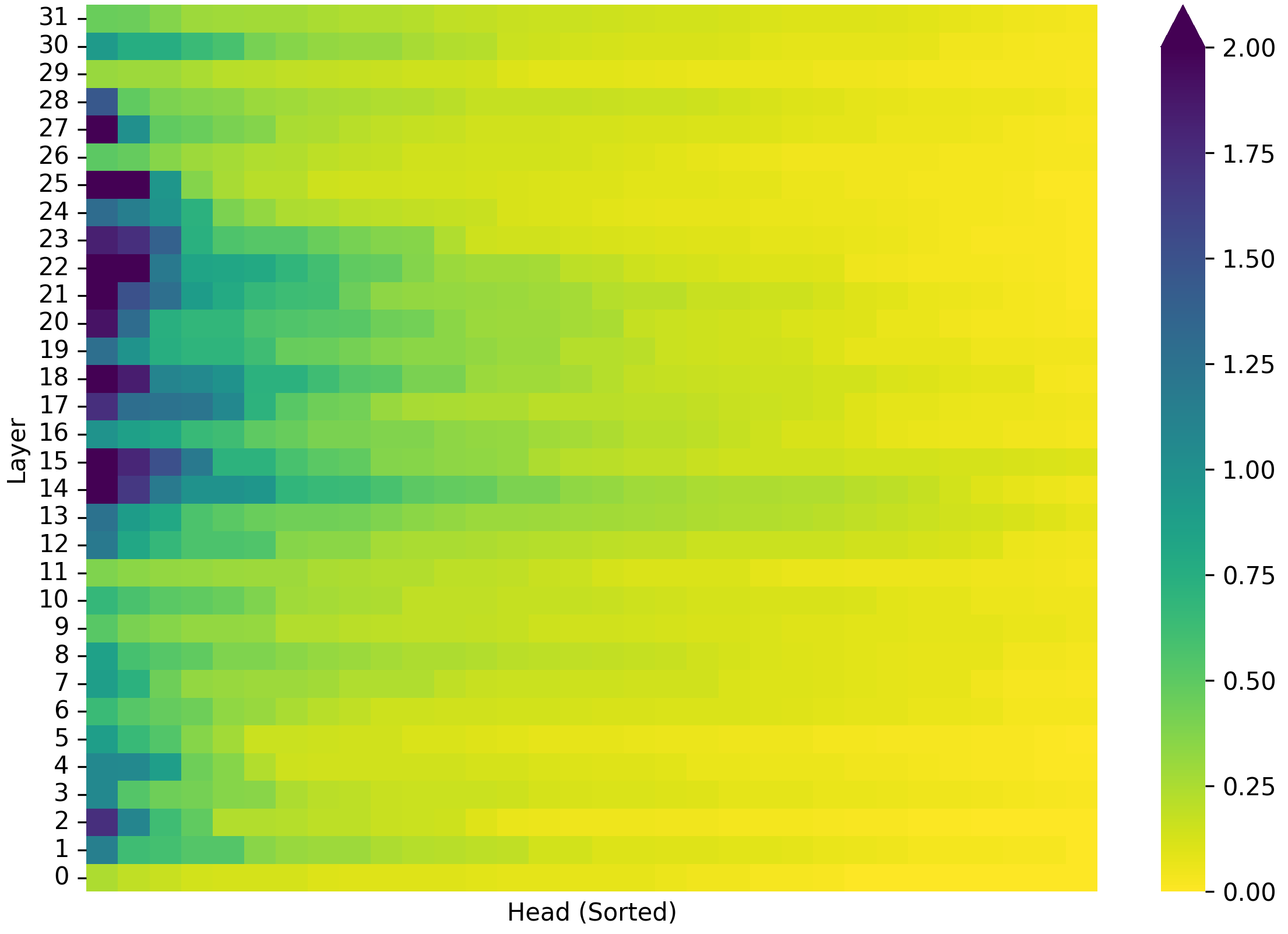}\hfill
\includegraphics[width=0.23\linewidth]{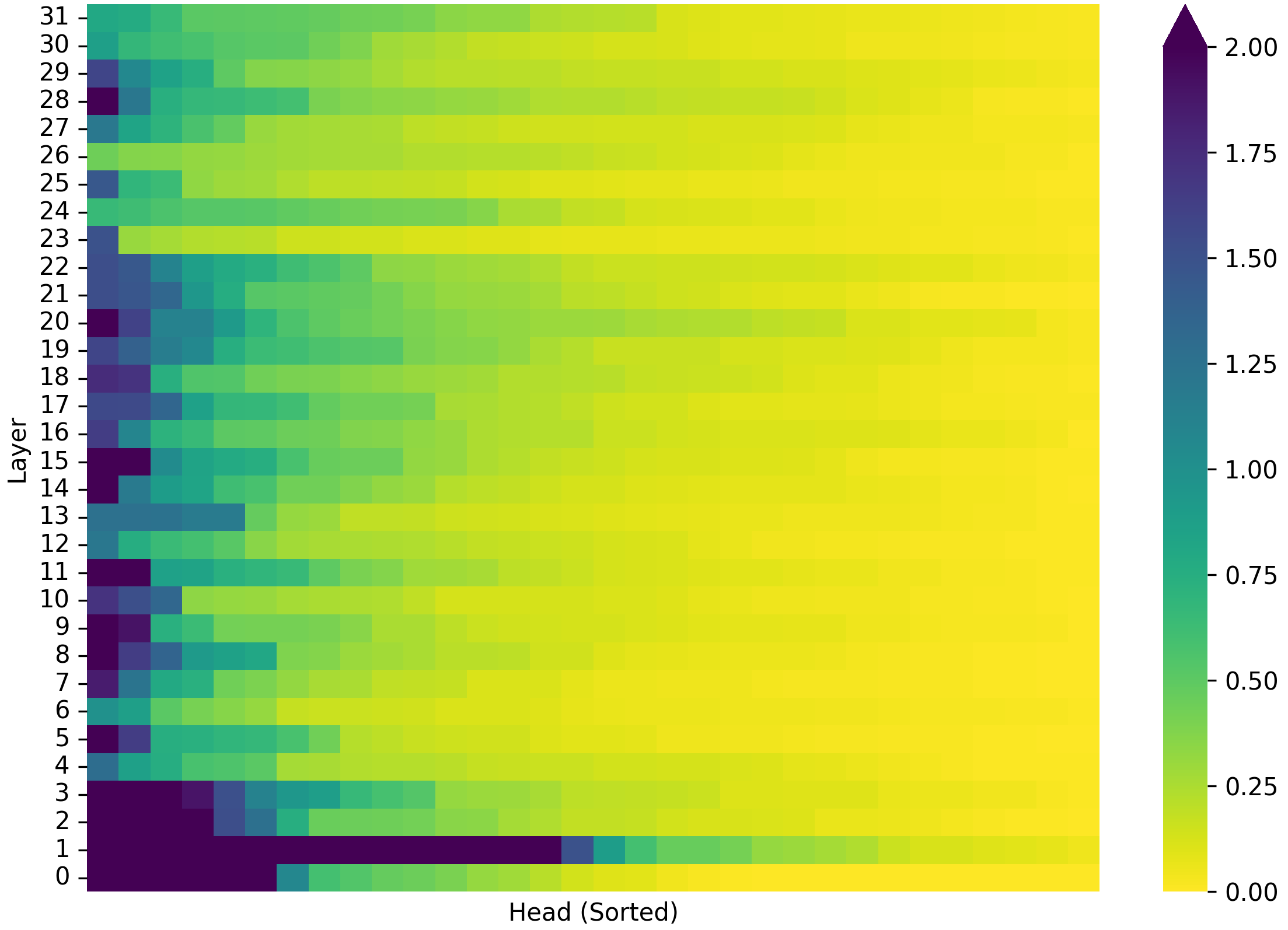}\hfill
\includegraphics[width=0.23\linewidth]{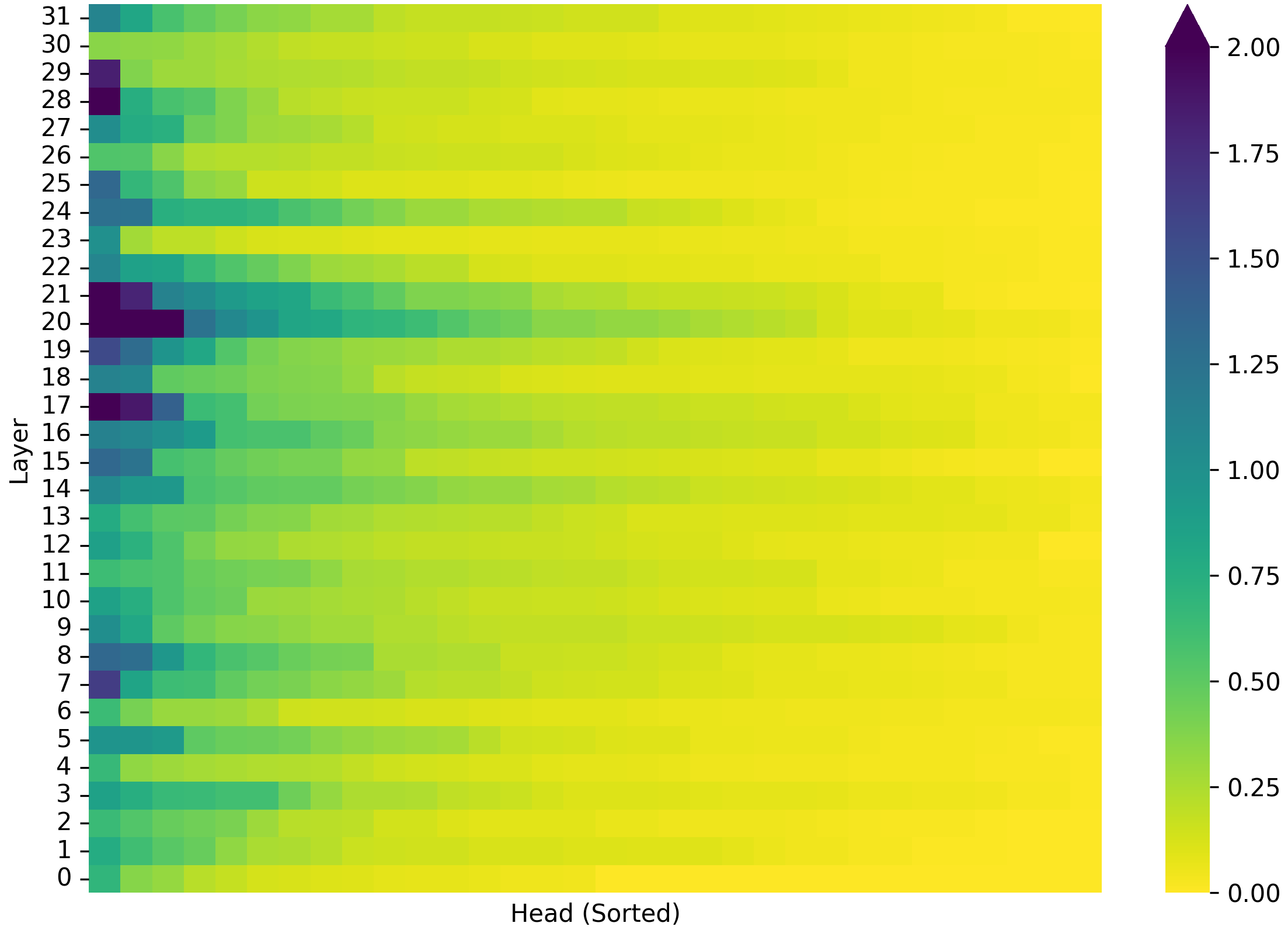}
\caption{\axis{theoretical.a.02}{applied.a.01}}
\label{fig:tp_theoretical}
\end{subfigure}

\caption{Variance-ratio heatmaps (Eq.~\ref{eq:score}) on \llama
(left two columns) and \mistral (right two columns) for
representative semantic axes. For each LLM, the left panel uses
the \textit{Last} token and the right panel uses the \textit{Mean}
over all tokens. \textit{Last}-token heatmaps concentrate signal
in early layers, while \textit{Mean}-over-tokens heatmaps localise
the highest variance ratios in middle-to-late layers, consistent
with where semantic representations emerge in decoder-only LLMs
\citep{marks2024geometrytruthemergentliner,
lavi-etal-2026-detecting, poulis2026testinglimitstruthdirections}.
Together with Table~\ref{tab:ablation}, this
motivates our choice of using \textit{Mean} over all tokens for extracting activations.}
\label{fig:ablation_position}
\end{figure*}

\paragraph{Token position.} Mean over all tokens outperforms
last token activations on Warmth for the two LLMs. On Competence the two converge at large $k$. Combined
with the layer location evidence in
Fig.~\ref{fig:ablation_position} --- where last token activations
concentrate signal in early layers, inconsistent with the
middle-to-late layers where semantic representations are known to
emerge in decoder-only LLMs
\citep{marks2024geometrytruthemergentliner,
lavi-etal-2026-detecting,
poulis2026testinglimitstruthdirections} --- this motivates our
choice of mean over all tokens in the main experiments.

\paragraph{Ensemble size.} Accuracy increases substantially with $k$
on Competence (from $\sim .50$ at $k = 1$ to $\sim .68$--$.69$ at
$k = 512$--$1024$) and modestly on Warmth (peaking around $k = 128$
before plateauing). This could indicate that informative attention heads related to Competence are distributed across a larger fraction of the model, whereas Warmth-related information is concentrated in fewer heads.

\paragraph{Selected configuration.} The
Listing template with $n = 30$ templates,  mean over all tokens, and $k \in [128, 512]$ is the best configuration in all
four (model, dimension) cells. We therefore use this configuration
in the main paper, fixing $k = 128$ as a single operating point
that balances Warmth (which peaks at $k = 128$) and Competence
(which keeps improving up to $k = 512$).

\begin{table*}[h]
\centering
\renewcommand{\arraystretch}{1.15}
\setlength{\tabcolsep}{4pt}
\resizebox{\textwidth}{!}{%
\begin{tabular}{lllll ccccccccc}
\toprule
\textbf{Model} & \textbf{Stereotype dimension} & \textbf{Template} & \textbf{Token} & \textbf{$n$} & \multicolumn{9}{c}{\textbf{Ensemble size $k$}} \\
\cmidrule(lr){6-14}
 & & & & & 1 & 8 & 16 & 32 & 64 & 128 & 256 & 512 & 1024 \\
\midrule
\multirow{16}{*}{\llama} & \multirow{8}{*}{Warmth (50)} & \multirow{4}{*}{Listing} & \multirow{2}{*}{Mean} & 15 & .50\tiny{(.16)} & .52\tiny{(.15)} & .52\tiny{(.13)} & .53\tiny{(.12)} & .54\tiny{(.13)} & \textbf{.55}\tiny{(.14)} & .54\tiny{(.13)} & .51\tiny{(.10)} & .51\tiny{(.10)} \\
 &  &  &  & 30 & .49\tiny{(.15)} & .52\tiny{(.15)} & .52\tiny{(.13)} & .53\tiny{(.12)} & .55\tiny{(.12)} & \underline{\textbf{.57}}\tiny{(.12)} & .54\tiny{(.13)} & .52\tiny{(.11)} & .52\tiny{(.11)} \\
 &  &  & \multirow{2}{*}{Last} & 15 & .48\tiny{(.12)} & .48\tiny{(.12)} & .49\tiny{(.13)} & .52\tiny{(.11)} & \textbf{.53}\tiny{(.12)} & .52\tiny{(.14)} & .53\tiny{(.14)} & .52\tiny{(.13)} & .52\tiny{(.12)} \\
 &  &  &  & 30 & .47\tiny{(.11)} & .48\tiny{(.11)} & .50\tiny{(.13)} & .52\tiny{(.12)} & \textbf{.53}\tiny{(.13)} & .53\tiny{(.13)} & .53\tiny{(.15)} & .52\tiny{(.13)} & .52\tiny{(.12)} \\
 &  & \multirow{4}{*}{Simple} & \multirow{2}{*}{Mean} & 15 & .52\tiny{(.15)} & .50\tiny{(.16)} & .52\tiny{(.14)} & .51\tiny{(.15)} & .52\tiny{(.15)} & .53\tiny{(.14)} & \textbf{.54}\tiny{(.14)} & .52\tiny{(.14)} & .50\tiny{(.13)} \\
 &  &  &  & 30 & .52\tiny{(.16)} & .50\tiny{(.13)} & .52\tiny{(.13)} & .52\tiny{(.14)} & .52\tiny{(.14)} & .53\tiny{(.13)} & \textbf{.54}\tiny{(.14)} & .52\tiny{(.13)} & .50\tiny{(.12)} \\
 &  &  & \multirow{2}{*}{Last} & 15 & .50\tiny{(.12)} & .49\tiny{(.13)} & .48\tiny{(.13)} & .51\tiny{(.11)} & .52\tiny{(.12)} & .51\tiny{(.12)} & .52\tiny{(.12)} & \textbf{.53}\tiny{(.13)} & .51\tiny{(.12)} \\
 &  &  &  & 30 & .47\tiny{(.11)} & .46\tiny{(.11)} & .48\tiny{(.10)} & .50\tiny{(.11)} & .51\tiny{(.12)} & .51\tiny{(.13)} & .53\tiny{(.12)} & \textbf{.54}\tiny{(.13)} & .51\tiny{(.12)} \\
\cmidrule(lr){2-14}
 & \multirow{8}{*}{Competence (42)} & \multirow{4}{*}{Listing} & \multirow{2}{*}{Mean} & 15 & .52\tiny{(.15)} & .55\tiny{(.11)} & .56\tiny{(.12)} & .57\tiny{(.14)} & .60\tiny{(.15)} & .62\tiny{(.17)} & .66\tiny{(.18)} & \textbf{.68}\tiny{(.18)} & .68\tiny{(.18)} \\
 &  &  &  & 30 & .53\tiny{(.14)} & .56\tiny{(.11)} & .56\tiny{(.12)} & .58\tiny{(.15)} & .61\tiny{(.17)} & .63\tiny{(.18)} & .67\tiny{(.18)} & \underline{\textbf{.69}}\tiny{(.19)} & .68\tiny{(.19)} \\
 &  &  & \multirow{2}{*}{Last} & 15 & .51\tiny{(.13)} & .52\tiny{(.14)} & .49\tiny{(.13)} & .54\tiny{(.16)} & .58\tiny{(.16)} & .62\tiny{(.20)} & .64\tiny{(.19)} & \textbf{.66}\tiny{(.19)} & .64\tiny{(.18)} \\
 &  &  &  & 30 & .51\tiny{(.14)} & .53\tiny{(.14)} & .52\tiny{(.14)} & .55\tiny{(.16)} & .57\tiny{(.16)} & .62\tiny{(.20)} & .64\tiny{(.21)} & .65\tiny{(.21)} & \textbf{.66}\tiny{(.19)} \\
 &  & \multirow{4}{*}{Simple} & \multirow{2}{*}{Mean} & 15 & .50\tiny{(.13)} & .53\tiny{(.10)} & .54\tiny{(.10)} & .55\tiny{(.14)} & .57\tiny{(.14)} & .60\tiny{(.17)} & .63\tiny{(.17)} & .65\tiny{(.18)} & \textbf{.67}\tiny{(.17)} \\
 &  &  &  & 30 & .52\tiny{(.12)} & .53\tiny{(.10)} & .54\tiny{(.11)} & .56\tiny{(.13)} & .59\tiny{(.14)} & .59\tiny{(.16)} & .62\tiny{(.17)} & .65\tiny{(.18)} & \textbf{.67}\tiny{(.18)} \\
 &  &  & \multirow{2}{*}{Last} & 15 & .53\tiny{(.12)} & .54\tiny{(.13)} & .54\tiny{(.15)} & .53\tiny{(.15)} & .56\tiny{(.16)} & .58\tiny{(.17)} & .61\tiny{(.17)} & \textbf{.64}\tiny{(.18)} & .64\tiny{(.19)} \\
 &  &  &  & 30 & .52\tiny{(.12)} & .53\tiny{(.12)} & .53\tiny{(.13)} & .52\tiny{(.15)} & .56\tiny{(.16)} & .59\tiny{(.18)} & .62\tiny{(.18)} & \textbf{.64}\tiny{(.19)} & .64\tiny{(.19)} \\
\midrule
\multirow{16}{*}{\mistral} & \multirow{8}{*}{Warmth (50)} & \multirow{4}{*}{Listing} & \multirow{2}{*}{Mean} & 15 & .52\tiny{(.14)} & .52\tiny{(.11)} & .52\tiny{(.11)} & \textbf{.55}\tiny{(.11)} & .54\tiny{(.12)} & .54\tiny{(.11)} & .54\tiny{(.12)} & .53\tiny{(.12)} & .52\tiny{(.12)} \\
 &  &  &  & 30 & .54\tiny{(.14)} & .52\tiny{(.10)} & .53\tiny{(.09)} & .55\tiny{(.11)} & .55\tiny{(.12)} & \underline{\textbf{.55}}\tiny{(.11)} & .55\tiny{(.11)} & .53\tiny{(.12)} & .52\tiny{(.12)} \\
 &  &  & \multirow{2}{*}{Last} & 15 & .47\tiny{(.10)} & .48\tiny{(.09)} & .48\tiny{(.10)} & .50\tiny{(.10)} & .51\tiny{(.11)} & .54\tiny{(.14)} & .54\tiny{(.12)} & \textbf{.55}\tiny{(.12)} & .53\tiny{(.12)} \\
 &  &  &  & 30 & .48\tiny{(.11)} & .48\tiny{(.09)} & .49\tiny{(.12)} & .50\tiny{(.09)} & .51\tiny{(.11)} & .53\tiny{(.13)} & .53\tiny{(.13)} & .54\tiny{(.12)} & \textbf{.54}\tiny{(.11)} \\
 &  & \multirow{4}{*}{Simple} & \multirow{2}{*}{Mean} & 15 & .53\tiny{(.12)} & .52\tiny{(.11)} & .52\tiny{(.11)} & .53\tiny{(.10)} & \textbf{.54}\tiny{(.11)} & .54\tiny{(.12)} & .53\tiny{(.11)} & .53\tiny{(.11)} & .52\tiny{(.13)} \\
 &  &  &  & 30 & .54\tiny{(.12)} & .54\tiny{(.11)} & .52\tiny{(.10)} & .54\tiny{(.11)} & .53\tiny{(.12)} & \textbf{.54}\tiny{(.11)} & .53\tiny{(.11)} & .53\tiny{(.11)} & .52\tiny{(.13)} \\
 &  &  & \multirow{2}{*}{Last} & 15 & .47\tiny{(.10)} & .49\tiny{(.10)} & .49\tiny{(.09)} & .50\tiny{(.09)} & .50\tiny{(.12)} & .49\tiny{(.11)} & .51\tiny{(.14)} & \textbf{.54}\tiny{(.13)} & .53\tiny{(.12)} \\
 &  &  &  & 30 & .47\tiny{(.11)} & .48\tiny{(.09)} & .48\tiny{(.09)} & .49\tiny{(.09)} & .49\tiny{(.11)} & .48\tiny{(.10)} & .52\tiny{(.14)} & \textbf{.54}\tiny{(.13)} & .54\tiny{(.14)} \\
\cmidrule(lr){2-14}
 & \multirow{8}{*}{Competence (42)} & \multirow{4}{*}{Listing} & \multirow{2}{*}{Mean} & 15 & .51\tiny{(.12)} & .58\tiny{(.13)} & .58\tiny{(.13)} & .60\tiny{(.13)} & .60\tiny{(.16)} & .62\tiny{(.15)} & .64\tiny{(.15)} & .66\tiny{(.16)} & \textbf{.67}\tiny{(.17)} \\
 &  &  &  & 30 & .49\tiny{(.12)} & .57\tiny{(.12)} & .58\tiny{(.14)} & .60\tiny{(.14)} & .60\tiny{(.16)} & .63\tiny{(.16)} & .65\tiny{(.16)} & .67\tiny{(.18)} & \underline{\textbf{.68}}\tiny{(.17)} \\
 &  &  & \multirow{2}{*}{Last} & 15 & .48\tiny{(.11)} & .49\tiny{(.11)} & .48\tiny{(.09)} & .49\tiny{(.10)} & .56\tiny{(.12)} & .62\tiny{(.14)} & .66\tiny{(.15)} & \textbf{.68}\tiny{(.17)} & .67\tiny{(.17)} \\
 &  &  &  & 30 & .51\tiny{(.12)} & .49\tiny{(.11)} & .48\tiny{(.11)} & .48\tiny{(.10)} & .56\tiny{(.12)} & .61\tiny{(.16)} & .67\tiny{(.16)} & \textbf{.68}\tiny{(.17)} & .67\tiny{(.18)} \\
 &  & \multirow{4}{*}{Simple} & \multirow{2}{*}{Mean} & 15 & .51\tiny{(.09)} & .54\tiny{(.10)} & .56\tiny{(.10)} & .58\tiny{(.12)} & .60\tiny{(.13)} & .61\tiny{(.13)} & .63\tiny{(.15)} & .65\tiny{(.15)} & \textbf{.66}\tiny{(.15)} \\
 &  &  &  & 30 & .50\tiny{(.12)} & .55\tiny{(.10)} & .57\tiny{(.12)} & .58\tiny{(.13)} & .59\tiny{(.14)} & .60\tiny{(.14)} & .64\tiny{(.15)} & \textbf{.66}\tiny{(.15)} & .66\tiny{(.15)} \\
 &  &  & \multirow{2}{*}{Last} & 15 & .51\tiny{(.12)} & .48\tiny{(.10)} & .49\tiny{(.13)} & .51\tiny{(.12)} & .53\tiny{(.12)} & .59\tiny{(.13)} & .62\tiny{(.14)} & \textbf{.65}\tiny{(.15)} & .64\tiny{(.15)} \\
 &  &  &  & 30 & .51\tiny{(.12)} & .49\tiny{(.10)} & .48\tiny{(.12)} & .50\tiny{(.11)} & .53\tiny{(.13)} & .59\tiny{(.13)} & .63\tiny{(.15)} & \textbf{.66}\tiny{(.16)} & .65\tiny{(.16)} \\
\bottomrule
\end{tabular}}
\caption{Position prediction accuracy (\S\ref{sec:rq2_positions}) of
\ours across all ablation configurations: template formulation
(Listing vs. Simple), token position ( Mean over all tokens 
vs.\ Last token), number of templates ($n \in \{15, 30\}$), and
ensemble size $k$.
Each cell reports the mean across axes with the standard deviation in parentheses. \textbf{Bold} marks the best $k$ within each row (per template/token/$n$ configuration); \underline{\textbf{bold and underlined}} marks the best cell over the entire block (across all configurations and $k$) for each (model, dimension). The best configuration is Listing + Mean over all tokens + $n{=}30$ in all four blocks. Counts next to stereotype dimensions are the number of semantic axes.}
\label{tab:ablation}
\end{table*}

\section{Case study datasets}
\label{app:casestudy_datasets}

\begin{table}[!t]
\centering
\small
\begin{tabular}{lrrr}
\toprule
\textbf{Stratum} & \textbf{Before} & \textbf{After} & \textbf{Sampled} \\
\midrule
Philosophy \& psychology & 206  & 81  & 34 \\
Unknown                  & 1487 & 429 & 33 \\
Technical domains        & 306  & 90  & 33 \\
\midrule
\textbf{Total}           & \textbf{1999} & \textbf{600} & \textbf{100} \\
\bottomrule
\end{tabular}
\caption{Stratum sizes of antonym pairs before filtering, after filtering, and the stratified sample quotas.}
\label{tab:domain_distribution}
\end{table}

\subsection{Social psychology surveys}
\label{app:psychology_surveys}
We  use ground truth labels that are well established across the social psychology literature, rather than ratings obtained from a limited annotator pool in a single computational study. To this end, we use the compilation of 
\citet{fraser-etal-2021-understanding}, who aggregated Warmth and Competence 
labels for 25 social group mentions (Table~\ref{tab:socail-groups-dataset1}). The labels are based on social psychology literature on gender and age 
stereotypes \citep{eckes2002paternalistic, cuddy2008warmth, fiske2018stereotype, 
glick2004bad}, occupational stereotypes \citep{brambilla2010effects, 
cuddy2011dynamics, fiske2010venus, losh2008methodological}, and ethnic and 
national group stereotypes \citep{lee2006not, fiske2006universal}.
 
 We operationalize Warmth as
Sociability $+$ Morality and Competence as Ability $+$ Agency.  We derive pole polarity from the labels in the stereotype dictionary of \citet{nicolas_2021}. For adjectives whose antonyms are not explicitly listed, we obtain them from WordNet, which is consistent with the stereotype dictionary because it adopts WordNet word senses. The full list of semantic axes for each category is provided in Table~\ref{tab:axes-dataset1}.

\begin{table}[ht]
\centering
\small
\setlength{\tabcolsep}{8pt}
\renewcommand{\arraystretch}{1.15}
\begin{tabular}{l c c}
\toprule
\textbf{Social group mention} & \textbf{Warmth} & \textbf{Competence} \\
\midrule
\multicolumn{3}{l}{\textit{Warm--Competent}} \\
Nurse                          & +1 & +1 \\
Psychologist                   & +1 & +1 \\
Researcher                     & +1 & +1 \\
\midrule
\multicolumn{3}{l}{\textit{Warm--Incompetent}} \\
Grandfather                    & +1 & -1 \\
Mommy                          & +1 & -1 \\
Mother                         & +1 & -1 \\
Schoolboy                      & +1 & -1 \\
Schoolgirl                     & +1 & -1 \\
\midrule
\multicolumn{3}{l}{\textit{Cold--Competent}} \\
Male                           & -1 & +1 \\
Gentleman                      & -1 & +1 \\
Japanese                       & -1 & +1 \\
Commander                      & -1 & +1 \\
Manager                        & -1 & +1 \\
Entrepreneur                   & -1 & +1 \\
Mathematician                  & -1 & +1 \\
Physicist                      & -1 & +1 \\
Chemist                        & -1 & +1 \\
Engineer                       & -1 & +1 \\
Software developer             & -1 & +1 \\
\midrule
\multicolumn{3}{l}{\textit{Cold--Incompetent}} \\
African                        & -1 & -1 \\
Ethiopian                      & -1 & -1 \\
Ghanaian                       & -1 & -1 \\
Eritrean                       & -1 & -1 \\
Hispanic                       & -1 & -1 \\
Arab                           & -1 & -1 \\
\bottomrule
\end{tabular}
\caption{Ground truth labels along the Warmth and Competence dimensions for the 25 social group mentions used in \secref{sec:rq2_positions}. The labels are drawn from prior 
work in social psychology collected by~\citet{fraser-etal-2021-understanding}.
Mentions are grouped by their SCM quadrant.}
\label{tab:socail-groups-dataset1}
\end{table}

\begin{table}[!t]
\centering
\footnotesize
\setlength{\tabcolsep}{3pt}
\renewcommand{\arraystretch}{1.15}
\begin{tabular}{@{}lcc@{}}
\toprule
\textbf{Category} & \textbf{Llama} & \textbf{Mistral} \\
\midrule
Flagged stereotypical ($p < 0.05$)     & 45 & 53 \\
\quad well-powered (pow.\ $\geq 0.80$) & 39 & 46 \\
\quad marginal (pow.\ $< 0.80$)        & 6  & 7  \\
\midrule
Not flagged ($p \geq 0.05$)            & 31 & 23 \\
\quad genuine nulls (pow.\ $< 0.30$)    & 13 & 7  \\
\quad underpowered (otherwise)         & 18 & 16 \\
\midrule
\textbf{Total applicable semantic axes}         & \textbf{76} & \textbf{76} \\
\bottomrule
\end{tabular}
\caption{Power analysis summary on the 76 applicable semantic axes. Llama
refers to \llama; Mistral refers to \mistral.}
\label{tab:power_summary}
\end{table}

\begin{table*}[h!]
\centering
\footnotesize
\setlength{\tabcolsep}{4pt}
\renewcommand{\arraystretch}{1.0}
\scalebox{0.85}{%
\begin{tabular}{
  l
  l
  l
  @{\hspace{10pt}}
  l
  l
  l
}
\toprule
\textbf{Dimension} & \textbf{Negative pole} & \textbf{Positive pole} &
\textbf{Dimension} & \textbf{Negative pole} & \textbf{Positive pole } \\
\midrule
\multirow{27}{*}{\textit{Morality}}
 & \texttt{immoral.a.01}            & \texttt{moral.a.01}
 & \multirow{23}{*}{\textit{Sociability }}
 & \texttt{unsociable.a.01}         & \texttt{sociable.a.01} \\

 & \texttt{insincere.a.01}          & \texttt{sincere.a.01}
 && \texttt{unfriendly.a.02}         & \texttt{friendly.a.01} \\

 & \texttt{dishonest.a.01}          & \texttt{honest.a.01}
 && \texttt{cold.a.02}               & \texttt{hot.a.03} \\

 & \texttt{egoistic.a.01}           & \texttt{altruistic.a.01}
 && \texttt{cool.a.04}               & \texttt{warm.a.02} \\

 & \texttt{selfish.a.01}            & \texttt{unselfish.a.01}
 && \texttt{unpleasant.a.01}         & \texttt{pleasant.a.01} \\

 & \texttt{hardhearted.a.01}        & \texttt{softhearted.a.01}
 && \texttt{disliked.a.01}           & \texttt{liked.a.01} \\

 & \texttt{disloyal.a.02}           & \texttt{loyal.a.01}
 && \texttt{insensitive.a.02}        & \texttt{sensitive.a.02} \\

 & \texttt{unfair.a.01}             & \texttt{fair.a.01}
 && \texttt{reserved.a.02}           & \texttt{unreserved.a.02} \\

 & \texttt{intolerant.a.01}         & \texttt{tolerant.a.01}
 && \texttt{unsympathetic.a.02}      & \texttt{sympathetic.a.04} \\

 & \texttt{evil.a.01}               & \texttt{good.a.03}
 && \texttt{unhelpful.a.01}          & \texttt{helpful.a.01} \\

 & \texttt{wicked.a.01}             & \texttt{virtuous.a.01}
 && \texttt{unsupportive.a.01}       & \texttt{supportive.a.01} \\

 & \texttt{unkind.a.01}             & \texttt{kind.a.01}
 && \texttt{impolite.a.01}           & \texttt{polite.a.01} \\

 & \texttt{wrong.a.02}              & \texttt{right.a.04}
 && \texttt{uncivil.a.01}            & \texttt{civil.a.02} \\

 & \texttt{corrupt.a.01}            & \texttt{incorrupt.a.01}
 && \texttt{unsocial.a.01}           & \texttt{social.a.02} \\

 & \texttt{guilty.a.01}             & \texttt{innocent.a.01}
 && \texttt{humorless.a.01}          & \texttt{humorous.a.01} \\

 & \texttt{hostile.a.01}            & \texttt{amicable.a.01}
 && \texttt{unpopular.a.01}          & \texttt{popular.a.01} \\

 & \texttt{inhumane.a.01}           & \texttt{humane.a.02}
 && \texttt{nasty.a.01}              & \texttt{nice.a.01} \\

 & \texttt{unfaithful.a.01}         & \texttt{faithful.a.01}
 && \texttt{tough.a.01}              & \texttt{tender.a.01} \\

 & \texttt{stingy.a.01}             & \texttt{generous.a.01}
 && \texttt{disagreeable.a.01}       & \texttt{agreeable.a.01} \\

 & \texttt{ill-natured.a.01}         & \texttt{good-natured.a.01}
 && \texttt{inhospitable.a.02}       & \texttt{hospitable.a.02} \\

 & \texttt{untruthful.a.01}         & \texttt{truthful.a.01}
 && \texttt{thoughtless.a.01}        & \texttt{thoughtful.a.02} \\

 & \texttt{uncompassionate.a.01}    & \texttt{compassionate.a.01}
 && \texttt{inconsiderate.a.01}     & \texttt{considerate.a.01} \\

 & \texttt{unforgiving.a.01}        & \texttt{forgiving.a.01}
 && \texttt{timid.a.01}     & \texttt{bold.a.01} \\

 & \texttt{resentful.a.01}          & \texttt{unresentful.a.01}
 &&              &  \\

 & \texttt{unreliable.a.02}         & \texttt{reliable.a.01}
 & \multirow{4}{*}{}
  &            & \\
 & \texttt{irresponsible.a.01}      & \texttt{responsible.a.01}
 
&&       &  \\
 & \texttt{prejudiced.a.02}         & \texttt{unprejudiced.a.01}

 &&         & \\
 &                                  &
 && & \\

\midrule
\multirow{19}{*}{\textit{Ability}}
 & \texttt{incompetent.a.02}        & \texttt{competent.a.01}
 & \multirow{23}{*}{\textit{Agency}}
 & \texttt{diffident.a.02}          & \texttt{confident.a.01} \\

 & \texttt{stupid.a.01}             & \texttt{smart.a.01}
 && \texttt{unassertive.a.01}        & \texttt{assertive.a.01} \\

 & \texttt{unintelligent.a.01}      & \texttt{intelligent.a.01}
 && \texttt{insecure.a.03}           & \texttt{secure.a.01} \\

 & \texttt{unable.a.01}             & \texttt{able.a.01}
 && \texttt{inactive.a.09}           & \texttt{active.a.05} \\

 & \texttt{unskilled.a.01}          & \texttt{skilled.a.01}
 && \texttt{dependent.a.01}          & \texttt{independent.a.01} \\

 & \texttt{uneducated.a.01}         & \texttt{educated.a.01}
 && \texttt{sporadic.a.01}           & \texttt{continual.a.01} \\

 & \texttt{irrational.a.01}         & \texttt{rational.a.01}
 && \texttt{unenterprising.a.01}     & \texttt{enterprising.a.01} \\

 & \texttt{uncreative.a.01}         & \texttt{creative.a.01}
 && \texttt{negligent.a.01}          & \texttt{diligent.a.02} \\

 & \texttt{incapable.a.01}          & \texttt{capable.a.01}
 && \texttt{lethargic.a.01}          & \texttt{energetic.a.01} \\

 & \texttt{impractical.a.01}        & \texttt{practical.a.01}
 && \texttt{unambitious.a.01}        & \texttt{ambitious.a.01} \\

 & \texttt{awkward.a.02}            & \texttt{graceful.a.01}
 && \texttt{undedicated.a.01}        & \texttt{dedicated.a.01} \\

 & \texttt{infelicitous.a.01}       & \texttt{felicitous.a.01}
 && \texttt{cautious.a.01}         & \texttt{incautious.a.01} \\

 & \texttt{foolish.a.01}            & \texttt{wise.a.01}
 && \texttt{irresolute.a.01}         & \texttt{resolute.a.01} \\

 & \texttt{uncritical.a.01}         & \texttt{critical.a.03}
 && \texttt{unadventurous.a.01}      & \texttt{adventurous.a.01} \\

 & \texttt{naive.a.01}              & \texttt{sophisticated.a.01}
 && \texttt{careless.a.01}           & \texttt{careful.a.01} \\

 & \texttt{undiscriminating.a.01}   & \texttt{discriminating.a.01}
 && \texttt{unmotivated.a.01}        & \texttt{motivated.a.01} \\

 & \texttt{maladroit.a.01}          & \texttt{adroit.a.01}
 && \texttt{spiritless.a.01}         & \texttt{spirited.a.01} \\

 & \texttt{illogical.a.01}          & \texttt{logical.a.01}
 && \texttt{troubled.a.01}         & \texttt{untroubled.a.01} \\

 & \texttt{unperceptive.a.01}       & \texttt{perceptive.a.02}
 && \texttt{subordinate.a.01}        & \texttt{dominant.a.01} \\

 &                                  &
 && \texttt{submissive.a.01}         & \texttt{domineering.a.01} \\

 &                                  &
 && \texttt{vulnerable.a.01}       & \texttt{invulnerable.a.01} \\

 &                                  &
 && \texttt{unaggressive.a.01}       & \texttt{aggressive.a.01} \\

 &                                  &
 && \texttt{docile.a.01}             & \texttt{stubborn.a.01} \\

\bottomrule
\end{tabular}
}
\caption{Semantic axes and their corresponding negative and positive poles used for the Warmth and Competence dimensions.
 We operationalize Warmth as
Sociability $+$ Morality and Competence as Ability $+$ Agency, following the stereotype dictionary of~\cite{nicolas_2021}. Semantic axes are grouped by stereotype dimension, and polarity assignments follow the labels provided in the stereotype dictionary. When antonyms required for constructing semantic axes were unavailable, they were obtained from WordNet, whose word senses are consistent with those used in the stereotype dictionary.}
\label{tab:axes-dataset1}
\end{table*}

\subsection{Our human annotations }

\subsubsection{Stratified Sampling}
\label{app:stratified}
Our goal was to obtain a stratified random sample of semantic axes from 
WordNet. We started from 1{,}999 pairs of antonymous adjective synsets from WordNet, where each pair consists of a positive and a negative pole. To assign each pair to a semantic domain, we used domain labels from BabelDomains~\cite{camacho_collados}, which defines 32 domains and propagates them to WordNet senses. We then applied four filtering steps. First, pairs whose two poles carried different BabelDomains labels were discarded. Second, pairs whose pole word was a quantifier or determiner (e.g., \textit{more}, \textit{fewer}, \textit{all}) were removed, since these do not constitute meaningful semantic axes. Third, pairs in which either pole definition contained a domain qualifier in parentheses, such as \textit{(of plants)} or \textit{(of mammals)}, were removed as they were not related to humans. Finally, we required both pole words to achieve a minimum Zipf score of 3.0 according to the \href{https://doi.org/10.5281/zenodo.7199437}{wordfreq} library, ensuring that both poles are sufficiently common in everyday language and thus meaningful for judging stereotype associations.

After filtering, 600 pairs formed the eligible pool. We drew a stratified random sample of 100 axes using a three-stratum design: \textit{Philosophy and psychology} (34), \textit{Unknown} (33), and \textit{Technical domains} (33), the last of which aggregates all remaining WordNet  domains, including Mathematics, Health and medicine, Physics and astronomy, Biology, Chemistry, and Linguistics. We separate \textit{Philosophy and psychology} from the other technical domains because it covers psychology and behavior and is thus most directly related to stereotypes, which are the primary interest of our analysis; isolating this stratum increases the chance of surfacing positive examples. Within each stratum, axes were sampled uniformly at random. Pairs whose poles admit no intermediate values (e.g., \vs{dead}{alive}, \vs{civilian}{military}) were replaced by another pair from the same stratum. Table~\ref{tab:domain_distribution} reports the stratum sizes and sample quotas. Table~\ref{tab:social-groups-dataset2} lists all 50 social group mentions used for human annotation, drawn from the social psychology literature on stereotypes. Tables~\ref{tab:power_llama_full_100} and~\ref{tab:power_mistral_full_100} report the 100 stratified semantic axes used in our experiments, together with the corresponding \ours outputs for \llama and \mistral.

\subsubsection{Annotation Process}
\label{app:annotation_protocol}
\begin{figure}[h]
    \centering
    \includegraphics[width=\linewidth]{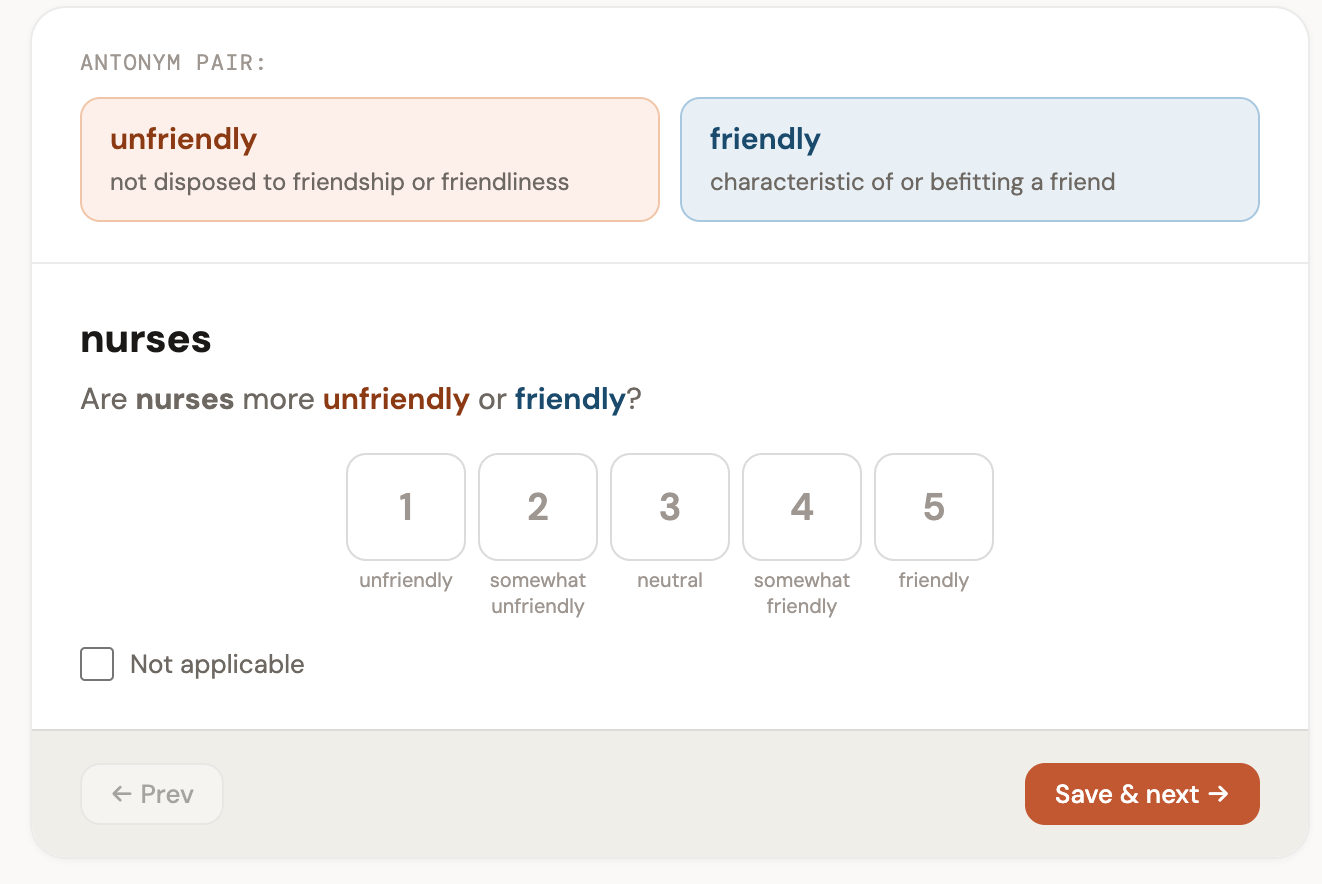}
    \caption{Screenshot of our annotation interface.}
    \label{fig:annotation_interface}
\end{figure}

\paragraph{Annotation task.}
We developed a web-based annotation interface (Fig.~\ref{fig:annotation_interface}) and provided annotators with the following instructions:

\begin{quote}
You will see antonym pairs (two opposite adjectives). For each pair, you will rate a social group on a 1--5 scale between the two adjectives, where 1 indicates the negative pole, 3 indicates neutrality, and 5 indicates the positive pole. If the pair does not apply to the social group, select \textit{Not applicable}. Please complete the annotation in one sitting without interruption.
\end{quote}

Rather than directly asking whether a semantic axis is stereotypical, which can be difficult to assess abstractly, we present a social group together with the semantic axis and ask annotators to rate the group comparatively along the semantic axis. For each item, annotators were also shown the WordNet definitions of both poles to disambiguate the senses.

\paragraph{Annotators and quality control.}
We initially recruited 11 volunteer annotators, who were unpaid. The annotators were informed of the sensitive nature of the task and consented to participation in the study. Each annotator rated 110 items: 100 items involving semantic axes paired with 50 social groups from Table~\ref{tab:social-groups-dataset2}, and 10 control items with unambiguous correct answers (e.g., \textit{``Are billionaires more poor or rich?''}) used for quality control. We excluded annotators who answered more than one control item incorrectly. This procedure removed 6 annotators, leaving 5 annotators in the final pool.

The annotators had diverse backgrounds across mother tongues (Chinese, Arabic, Romanian, Spanish, English, Farsi, and German), education levels (bachelor's, master's, and PhD), and disciplines (STEM and social sciences). Most reported English proficiency at the C1/C2 level. This diversity was intentional, ensuring that the annotations reflected a broad range of perspectives.

\paragraph{Annotation aggregation.}
\label{app:annotation_aggregation}
For each semantic axis, we first determine its applicability. A semantic axis is labeled \textit{Not applicable} if a majority of annotators vote that the axis is not applicable. After excluding these axes, 76 semantic axes remain. For the remaining axes, we map ratings into three categories: below 3, equal to 3, and above 3. If a majority of annotators who provided a rating fall within one of these categories after aggregation, we label the semantic axis as \textit{stereotypical}; otherwise, we label it as \textit{non-stereotypical}. We measure inter-annotator agreement using Fleiss' kappa, obtaining Fleiss'
$\kappa = 0.375$, which indicates fair agreement. 

\begin{table*}[ht]
\centering
\begin{tabular}{r l r @{\hspace{8pt}} r l r}
\toprule
\textbf{\#} & \textbf{Social group} & \textbf{Zipf} 
& \textbf{\#} & \textbf{Social group} & \textbf{Zipf} \\
\midrule
1  & women                & 5.57 & 26 & refugees                & 4.26 \\
2  & men                  & 5.51 & 27 & elderly people          & 4.15 \\
3  & White people         & 5.51 & 28 & nurses                  & 4.14 \\
4  & Black people         & 5.46 & 29 & conservatives           & 4.14 \\
5  & working class people & 5.36 & 30 & liberals                & 4.13 \\
6  & middle class people  & 5.18 & 31 & Hindu people            & 4.00 \\
7  & poor people          & 5.09 & 32 & blue-collar workers     & 4.00 \\
8  & career women         & 5.07 & 33 & Hispanic people         & 3.97 \\
9  & rich people          & 4.91 & 34 & professors              & 3.92 \\
10 & gay people           & 4.90 & 35 & transgender people      & 3.91 \\
11 & Christian people     & 4.88 & 36 & unemployed people       & 3.90 \\
12 & traditional women    & 4.84 & 37 & welfare recipients      & 3.82 \\
13 & upper class people   & 4.72 & 38 & believers               & 3.66 \\
14 & Asian people         & 4.69 & 39 & feminists               & 3.61 \\
15 & Muslim people        & 4.67 & 40 & drug addicts            & 3.50 \\
16 & Jewish people        & 4.61 & 41 & intellectuals           & 3.49 \\
17 & doctors              & 4.55 & 42 & undocumented immigrants & 3.44 \\
18 & Democrats            & 4.54 & 43 & billionaires            & 3.44 \\
19 & Republicans          & 4.50 & 44 & LGBTQ people            & 3.41 \\
20 & scientists           & 4.50 & 45 & housewives              & 3.40 \\
21 & managers             & 4.47 & 46 & CEOs                    & 3.39 \\
22 & engineers            & 4.41 & 47 & atheists                & 3.34 \\
23 & immigrants           & 4.37 & 48 & uneducated people       & 3.24 \\
24 & disabled people      & 4.29 & 49 & skeptics                & 3.07 \\
25 & homeless people      & 4.26 & 50 & agnostics               & 2.38 \\
\bottomrule
\end{tabular}
\caption{The list of 50 social group mentions used in \secref{sec:rq2_axes}, sorted by Zipf frequency. }
\label{tab:social-groups-dataset2}
\end{table*}

\begin{table*}[ht]
\centering
\begin{tabular}{r l r @{\hspace{8pt}} r l r}
\toprule
\textbf{\#} & \textbf{Random phrases} & \textbf{Zipf} 
& \textbf{\#} & \textbf{Random phrases} & \textbf{Zipf} \\
\midrule
1  & country house        & 5.28 & 26 & block and tackle        & 4.08 \\
2  & record               & 5.21 & 27 & toy soldier             & 4.07 \\
3  & lead                 & 5.20 & 28 & basket                  & 4.07 \\
4  & church               & 5.16 & 29 & span                    & 4.07 \\
5  & saw set              & 5.15 & 30 & portal                  & 4.00 \\
6  & local street         & 5.05 & 31 & outlet                  & 4.00 \\
7  & music stand          & 4.99 & 32 & cargo vessel            & 3.93 \\
8  & trip                 & 4.90 & 33 & coast lily              & 3.90 \\
9  & push                 & 4.82 & 34 & atomic bomb             & 3.89 \\
10 & stage door           & 4.79 & 35 & lime                    & 3.86 \\
11 & figure         & 4.74 & 36 & ginger nut              & 3.68 \\
12 & television room      & 4.69 & 37 & cinnamon bread          & 3.64 \\
13 & nick                 & 4.59 & 38 & rotary engine           & 3.59 \\
14 & excuse               & 4.56 & 39 & tor                     & 3.57 \\
15 & coach horse          & 4.55 & 40 & fairy lantern           & 3.48 \\
16 & main yard            & 4.49 & 41 & prism                   & 3.44 \\
17 & tank ship            & 4.41 & 42 & reflex camera           & 3.42 \\
18 & flying bomb          & 4.35 & 43 & bandwagon               & 3.26 \\
19 & lower deck           & 4.33 & 44 & cheesecake              & 3.22 \\
20 & railroad             & 4.32 & 45 & capacitor               & 3.17 \\
21 & carrier              & 4.29 & 46 & paper mulberry          & 3.15 \\
22 & instrument           & 4.26 & 47 & auditory cortex         & 3.09 \\
23 & double bass          & 4.21 & 48 & toner                   & 3.07 \\
24 & dinner theater       & 4.21 & 49 & tarmac                  & 3.04 \\
25 & delayed action       & 4.16 & 50 & footpath                    & 2.99 \\
\bottomrule
\end{tabular}
\caption{The 50 random nouns and noun phrases sampled from WordNet and used in \secref{sec:rq2_axes}. These were selected to match the Zipf frequency distribution of the social group mentions in Table~\ref{tab:social-groups-dataset2}.}
\label{tab:random-nouns}
\end{table*}

\section{Power Analysis}
\label{app:power_analysis}


We estimate the reliability of the Kolmogorov--Smirnov test conducted in \S\ref{sec:hypothesis} by Monte Carlo simulation~\citep{mooney1997}.
We draw 1{,}000 bootstrap resamples of size $n_{\mathcal{C}} = n_{\mathcal{C}'} = 50$ 
from the observed projections and report the empirical power against 
the observed effect at the current sample size, with significance 
level $\alpha = 0.05$ and target power $0.80$~\citep{cohen1988statistical}.

We classify semantic axes into three groups: \textit{well-powered positives}
(flagged stereotypical, power $\geq 0.80$); \textit{genuine nulls}
(not flagged, power $< 0.30$, where the test would reliably detect
even moderate effects but finds none); and \textit{underpowered}
(not flagged, power between $0.30$ and $0.80$, where
non-significance is inconclusive)~\citep{ozccomak2013comparison}. 
Table~\ref{tab:power_summary} reports the counts in each category 
for \llama and \mistral. Per-axis power analysis of the Kolmogorov--Smirnov test for \ours instantiated with the both LLMs are reported in Tables~\ref{tab:power_llama_full_100} 
and~\ref{tab:power_mistral_full_100}.

\begin{table*}[ht]
\centering
\scriptsize
\setlength{\tabcolsep}{2.2pt}
\renewcommand{\arraystretch}{1.02}
\begin{tabular}{@{}lcccccc@{\hspace{4pt}}|@{\hspace{4pt}}lcccccc@{}}
\toprule
\textbf{Axis (Neg./Pos.)} & $D$ & $p$ & Pow. & $n_{80}$ & Cat. & Hum. &
\textbf{Axis (Neg./Pos.)} & $D$ & $p$ & Pow. & $n_{80}$ & Cat. & Hum. \\
\midrule
\texttt{\tiny insensitive.a.02/sensitive.a.02} & .72 & \textless.001 & 1.00 & 10 & \textsc{wp} & \ding{51} & \texttt{\tiny light-duty.a.01/heavy-duty.a.01} & .16 & .549 & .14 & 141 & \textsc{gn} & \ding{55} \\
\texttt{\tiny close.a.02/distant.a.02} & .64 & \textless.001 & 1.00 & 12 & \textsc{wp} & \ding{51} & \texttt{\tiny unworthy.a.01/worthy.a.01} & .16 & .549 & .21 & 138 & \textsc{gn} & \ding{51} \\
\texttt{\tiny incompetent.a.01/competent.a.03} & .64 & \textless.001 & 1.00 & 12 & \textsc{wp} & \ding{51} & \texttt{\tiny hard.a.03/soft.a.01} & .14 & .717 & .17 & 218 & \textsc{gn} & \ding{55} \\
\texttt{\tiny indecisive.a.02/decisive.a.01} & .64 & \textless.001 & 1.00 & 12 & \textsc{wp} & \ding{51} & \texttt{\tiny harmless.a.01/harmful.a.01} & .14 & .717 & .20 & 165 & \textsc{gn} & \ding{51} \\
\texttt{\tiny unreasonable.a.01/reasonable.a.01} & .62 & \textless.001 & 1.00 & 12 & \textsc{wp} & \ding{51} & \texttt{\tiny immoral.a.01/moral.a.01} & .14 & .717 & .12 & 189 & \textsc{gn} & \ding{51} \\
\texttt{\tiny dissimilar.a.01/similar.a.01} & .62 & \textless.001 & 1.00 & 11 & \textsc{wp} & \ding{51} & \texttt{\tiny negative.a.05/positive.a.08} & .14 & .717 & .11 & 194 & \textsc{gn} & \ding{51} \\
\texttt{\tiny straight.a.08/curved.a.01} & .62 & \textless.001 & 1.00 & 12 & \textsc{wp} & \ding{55} & \texttt{\tiny tame.a.02/wild.a.01} & .12 & .869 & .08 & 288 & \textsc{gn} & \ding{55} \\
\texttt{\tiny cowardly.a.01/brave.a.01} & .56 & \textless.001 & 1.00 & 12 & \textsc{wp} & \ding{51} & \texttt{\tiny ill.a.01/well.a.01} & .10 & .967 & .10 & 321 & \textsc{gn} & \ding{55} \\
\texttt{\tiny untrained.a.01/trained.a.01} & .56 & \textless.001 & 1.00 & 16 & \textsc{wp} & \ding{51} & \texttt{\tiny unambitious.a.01/ambitious.a.01} & .26 & .068 & .77 & 58 & \textsc{up} & \ding{51} \\
\texttt{\tiny timid.a.01/bold.a.01} & .54 & \textless.001 & 1.00 & 15 & \textsc{wp} & \ding{55} & \texttt{\tiny lower-class.a.01/upper-class.a.01} & .26 & .068 & .66 & 59 & \textsc{up} & \ding{51} \\
\texttt{\tiny illiterate.a.03/literate.a.02} & .54 & \textless.001 & 1.00 & 13 & \textsc{wp} & \ding{51} & \texttt{\tiny careless.a.01/careful.a.01} & .26 & .068 & .62 & 67 & \textsc{up} & \ding{55} \\
\texttt{\tiny imperfect.a.01/perfect.a.01} & .54 & \textless.001 & 1.00 & 12 & \textsc{wp} & \ding{51} & \texttt{\tiny unfaithful.a.01/faithful.a.01} & .26 & .068 & .74 & 58 & \textsc{up} & \ding{55} \\
\texttt{\tiny unwelcome.a.01/welcome.a.01} & .54 & \textless.001 & 1.00 & 13 & \textsc{wp} & \ding{51} & \texttt{\tiny uncompromising.a.01/compromising.a.01} & .26 & .068 & .71 & 57 & \textsc{up} & \ding{51} \\
\texttt{\tiny unwilling.a.01/willing.a.01} & .54 & \textless.001 & 1.00 & 16 & \textsc{wp} & \ding{51} & \texttt{\tiny pessimistic.a.01/optimistic.a.01} & .26 & .068 & .63 & 67 & \textsc{up} & \ding{55} \\
\texttt{\tiny dull.a.09/sharp.a.09} & .50 & \textless.001 & 1.00 & 16 & \textsc{wp} & \ding{55} & \texttt{\tiny unsettled.a.01/settled.a.01} & .26 & .068 & .63 & 67 & \textsc{up} & \ding{51} \\
\texttt{\tiny poor.a.02/rich.a.01} & .50 & \textless.001 & 1.00 & 17 & \textsc{wp} & \ding{55} & \texttt{\tiny inconsiderate.a.01/considerate.a.01} & .24 & .112 & .49 & 83 & \textsc{up} & \ding{51} \\
\texttt{\tiny stupid.a.01/smart.a.01} & .48 & \textless.001 & 1.00 & 17 & \textsc{wp} & \ding{51} & \texttt{\tiny negligent.a.01/diligent.a.02} & .24 & .112 & .45 & 89 & \textsc{up} & \ding{51} \\
\texttt{\tiny ungrateful.a.01/grateful.a.01} & .46 & \textless.001 & 1.00 & 20 & \textsc{wp} & \ding{51} & \texttt{\tiny masculine.a.01/feminine.a.02} & .24 & .112 & .65 & 60 & \textsc{up} & \ding{51} \\
\texttt{\tiny irregular.a.01/regular.a.01} & .44 & \textless.001 & 1.00 & 21 & \textsc{wp} & \ding{51} & \texttt{\tiny unsatisfactory.a.01/satisfactory.a.01} & .24 & .112 & .67 & 66 & \textsc{up} & \ding{51} \\
\texttt{\tiny depressing.a.01/cheerful.a.01} & .42 & \textless.001 & .99 & 26 & \textsc{wp} & \ding{51} & \texttt{\tiny ill-advised.a.01/well-advised.a.01} & .22 & .179 & .48 & 90 & \textsc{up} & \ding{51} \\
\texttt{\tiny unconventional.a.02/conventional.a.01} & .42 & \textless.001 & 1.00 & 21 & \textsc{wp} & \ding{55} & \texttt{\tiny slow.a.01/fast.a.01} & .22 & .179 & .50 & 90 & \textsc{up} & \ding{55} \\
\texttt{\tiny inhumane.a.01/humane.a.02} & .42 & \textless.001 & 1.00 & 20 & \textsc{wp} & \ding{55} & \texttt{\tiny quiet.a.02/noisy.a.01} & .22 & .179 & .38 & 102 & \textsc{up} & \ding{55} \\
\texttt{\tiny powerless.a.01/powerful.a.01} & .42 & \textless.001 & 1.00 & 21 & \textsc{wp} & \ding{51} & \texttt{\tiny intolerant.a.01/tolerant.a.01} & .22 & .179 & .44 & 90 & \textsc{up} & \ding{55} \\
\texttt{\tiny impure.a.02/pure.a.06} & .42 & \textless.001 & .99 & 25 & \textsc{wp} & \ding{51} & \texttt{\tiny unfortunate.a.01/fortunate.a.01} & .20 & .272 & .44 & 94 & \textsc{up} & \ding{51} \\
\texttt{\tiny inferior.a.01/superior.a.02} & .42 & \textless.001 & .98 & 26 & \textsc{wp} & \ding{51} & \texttt{\tiny unfriendly.a.02/friendly.a.01} & .18 & .396 & .32 & 106 & \textsc{up} & \ding{51} \\
\texttt{\tiny unattractive.a.01/attractive.a.01} & .40 & \textless.001 & 1.00 & 21 & \textsc{wp} & \ding{55} & \texttt{\tiny illogical.a.01/logical.a.01} & .18 & .396 & .38 & 96 & \textsc{up} & \ding{51} \\
\texttt{\tiny destructive.a.01/constructive.a.01} & .38 & .001 & .96 & 35 & \textsc{wp} & \ding{51} & \texttt{\tiny lost.a.03/saved.a.01} & .54 & \textless.001 & 1.00 & 16 & \textsc{wp} & -- \\
\texttt{\tiny wrong.a.02/right.a.04} & .38 & .001 & .94 & 35 & \textsc{wp} & \ding{51} & \texttt{\tiny green.a.03/ripe.a.01} & .50 & \textless.001 & 1.00 & 17 & \textsc{wp} & -- \\
\texttt{\tiny unemployed.a.01/employed.a.01} & .36 & .003 & .97 & 34 & \textsc{wp} & \ding{51} & \texttt{\tiny small.a.01/large.a.01} & .44 & \textless.001 & .99 & 26 & \textsc{wp} & -- \\
\texttt{\tiny dishonest.a.01/honest.a.01} & .36 & .003 & .92 & 37 & \textsc{wp} & \ding{51} & \texttt{\tiny destroyed.a.01/preserved.a.02} & .42 & \textless.001 & .98 & 27 & \textsc{wp} & -- \\
\texttt{\tiny mild.a.01/intense.a.01} & .36 & .003 & .94 & 35 & \textsc{wp} & \ding{51} & \texttt{\tiny rough.a.01/smooth.a.01} & .40 & \textless.001 & .96 & 34 & \textsc{wp} & -- \\
\texttt{\tiny dull.a.02/bright.a.01} & .34 & .006 & .93 & 37 & \textsc{wp} & \ding{51} & \texttt{\tiny worn.a.01/new.a.06} & .40 & \textless.001 & .95 & 35 & \textsc{wp} & -- \\
\texttt{\tiny narrow-minded.a.02/broad-minded.a.02} & .34 & .006 & .91 & 37 & \textsc{wp} & \ding{51} & \texttt{\tiny light.a.05/heavy.a.04} & .38 & .001 & .95 & 34 & \textsc{wp} & -- \\
\texttt{\tiny subordinate.a.01/dominant.a.01} & .34 & .006 & .89 & 43 & \textsc{wp} & \ding{51} & \texttt{\tiny asymmetrical.a.01/symmetrical.a.01} & .36 & .003 & .93 & 36 & \textsc{wp} & -- \\
\texttt{\tiny retarded.a.01/precocious.a.01} & .32 & .011 & .85 & 44 & \textsc{wp} & \ding{51} & \texttt{\tiny painless.a.02/painful.a.01} & .32 & .011 & .77 & 51 & \textsc{m} & -- \\
\texttt{\tiny insane.a.01/sane.a.01} & .32 & .011 & .89 & 43 & \textsc{wp} & \ding{55} & \texttt{\tiny acidic.a.01/alkaline.a.01} & .30 & .022 & .68 & 57 & \textsc{m} & -- \\
\texttt{\tiny foolish.a.01/wise.a.01} & .32 & .011 & .82 & 49 & \textsc{wp} & \ding{55} & \texttt{\tiny insufficient.a.01/sufficient.a.01} & .30 & .022 & .78 & 51 & \textsc{m} & -- \\
\texttt{\tiny unimportant.a.01/important.a.01} & .30 & .022 & .82 & 51 & \textsc{wp} & \ding{55} & \texttt{\tiny unspecified.a.01/specified.a.01} & .28 & .039 & .78 & 53 & \textsc{m} & -- \\
\texttt{\tiny segregated.a.01/integrated.a.03} & .30 & .022 & .84 & 45 & \textsc{wp} & \ding{51} & \texttt{\tiny horizontal.a.01/vertical.a.01} & .18 & .396 & .28 & 136 & \textsc{gn} & -- \\
\texttt{\tiny loose.a.01/compact.a.01} & .30 & .022 & .71 & 60 & \textsc{m} & \ding{51} & \texttt{\tiny dirty.a.01/clean.a.01} & .16 & .549 & .12 & 179 & \textsc{gn} & -- \\
\texttt{\tiny left.a.04/right.a.07} & .30 & .022 & .78 & 51 & \textsc{m} & \ding{55} & \texttt{\tiny proximal.a.01/distal.a.01} & .16 & .549 & .18 & 209 & \textsc{gn} & -- \\
\texttt{\tiny tasteless.a.02/tasteful.a.01} & .30 & .022 & .76 & 52 & \textsc{m} & \ding{55} & \texttt{\tiny south.a.01/north.a.01} & .12 & .869 & .14 & 286 & \textsc{gn} & -- \\
\texttt{\tiny indirect.a.04/direct.a.03} & .28 & .039 & .74 & 58 & \textsc{m} & \ding{51} & \texttt{\tiny earthly.a.01/heavenly.a.03} & .10 & .967 & .08 & 386 & \textsc{gn} & -- \\
\texttt{\tiny stingy.a.01/generous.a.01} & .28 & .039 & .78 & 52 & \textsc{m} & \ding{55} & \texttt{\tiny downtown.a.01/uptown.a.01} & .10 & .967 & .08 & 358 & \textsc{gn} & -- \\
\texttt{\tiny humble.a.02/proud.a.01} & .28 & .039 & .67 & 61 & \textsc{m} & \ding{51} & \texttt{\tiny difficult.a.01/easy.a.01} & .26 & .068 & .72 & 63 & \textsc{up} & -- \\
\texttt{\tiny following.a.03/leading.a.03} & .20 & .272 & .29 & 140 & \textsc{gn} & \ding{51} & \texttt{\tiny theoretical.a.02/applied.a.01} & .24 & .112 & .59 & 73 & \textsc{up} & -- \\
\texttt{\tiny uncomfortable.a.01/comfortable.a.02} & .18 & .396 & .27 & 135 & \textsc{gn} & \ding{51} & \texttt{\tiny sweet.a.01/sour.a.02} & .24 & .112 & .56 & 90 & \textsc{up} & -- \\
\texttt{\tiny frivolous.a.01/serious.a.01} & .18 & .396 & .18 & 140 & \textsc{gn} & \ding{51} & \texttt{\tiny short.a.06/long.a.05} & .22 & .179 & .50 & 92 & \textsc{up} & -- \\
\texttt{\tiny weak.a.01/strong.a.01} & .18 & .396 & .27 & 125 & \textsc{gn} & \ding{51} & \texttt{\tiny central.a.02/peripheral.a.01} & .20 & .272 & .30 & 124 & \textsc{up} & -- \\
\texttt{\tiny sterile.a.01/fertile.a.01} & .16 & .549 & .16 & 142 & \textsc{gn} & \ding{51} & \texttt{\tiny intracellular.a.01/extracellular.a.01} & .20 & .272 & .36 & 114 & \textsc{up} & -- \\
\bottomrule
\end{tabular}
\caption{Per-axis power analysis for \ours instantiated with \llama across all 100 sampled semantic axes. $D$: two-sample Kolmogorov--Smirnov statistic (\secref{sec:hypothesis}); $p$: $p$-value; Pow.: empirical power at $n_{\mathcal{C}} = n_{\mathcal{C}'} = 50$ from 1{,}000 bootstrap resamples; $n_{80}$: estimated sample size per group to reach power $0.80$; Cat.: \textsc{wp} well-powered positive, \textsc{m} marginal, \textsc{gn} genuine null, \textsc{up} underpowered; Hum.: human annotators judge the semantic axis stereotypical (\ding{51}), non-stereotypical (\ding{55}), or not applicable (--).}
\label{tab:power_llama_full_100}
\end{table*}

\begin{table*}[ht]
\centering
\scriptsize
\setlength{\tabcolsep}{2.2pt}
\renewcommand{\arraystretch}{1.02}
\begin{tabular}{@{}lcccccc@{\hspace{4pt}}|@{\hspace{4pt}}lcccccc@{}}
\toprule
\textbf{Axis (Neg./Pos.)} & $D$ & $p$ & Pow. & $n_{80}$ & Cat. & Hum. &
\textbf{Axis (Neg./Pos.)} & $D$ & $p$ & Pow. & $n_{80}$ & Cat. & Hum. \\
\midrule
\texttt{\tiny imperfect.a.01/perfect.a.01} & .68 & \textless.001 & 1.00 & 11 & \textsc{wp} & \ding{51} & \texttt{\tiny subordinate.a.01/dominant.a.01} & .28 & .039 & .72 & 58 & \textsc{m} & \ding{51} \\
\texttt{\tiny dull.a.09/sharp.a.09} & .62 & \textless.001 & 1.00 & 12 & \textsc{wp} & \ding{55} & \texttt{\tiny masculine.a.01/feminine.a.02} & .28 & .039 & .70 & 60 & \textsc{m} & \ding{51} \\
\texttt{\tiny narrow-minded.a.02/broad-minded.a.02} & .60 & \textless.001 & 1.00 & 12 & \textsc{wp} & \ding{51} & \texttt{\tiny impure.a.02/pure.a.06} & .28 & .039 & .73 & 56 & \textsc{m} & \ding{51} \\
\texttt{\tiny close.a.02/distant.a.02} & .58 & \textless.001 & 1.00 & 12 & \textsc{wp} & \ding{51} & \texttt{\tiny sterile.a.01/fertile.a.01} & .20 & .272 & .23 & 118 & \textsc{gn} & \ding{51} \\
\texttt{\tiny dissimilar.a.01/similar.a.01} & .54 & \textless.001 & 1.00 & 16 & \textsc{wp} & \ding{51} & \texttt{\tiny immoral.a.01/moral.a.01} & .20 & .272 & .26 & 116 & \textsc{gn} & \ding{51} \\
\texttt{\tiny unwelcome.a.01/welcome.a.01} & .54 & \textless.001 & 1.00 & 15 & \textsc{wp} & \ding{51} & \texttt{\tiny weak.a.01/strong.a.01} & .20 & .272 & .26 & 118 & \textsc{gn} & \ding{51} \\
\texttt{\tiny timid.a.01/bold.a.01} & .52 & \textless.001 & 1.00 & 17 & \textsc{wp} & \ding{55} & \texttt{\tiny uncompromising.a.01/compromising.a.01} & .16 & .549 & .25 & 142 & \textsc{gn} & \ding{51} \\
\texttt{\tiny incompetent.a.01/competent.a.03} & .52 & \textless.001 & 1.00 & 16 & \textsc{wp} & \ding{51} & \texttt{\tiny humble.a.02/proud.a.01} & .16 & .549 & .11 & 168 & \textsc{gn} & \ding{51} \\
\texttt{\tiny unattractive.a.01/attractive.a.01} & .50 & \textless.001 & 1.00 & 17 & \textsc{wp} & \ding{55} & \texttt{\tiny hard.a.03/soft.a.01} & .14 & .717 & .08 & 241 & \textsc{gn} & \ding{55} \\
\texttt{\tiny indecisive.a.02/decisive.a.01} & .50 & \textless.001 & 1.00 & 17 & \textsc{wp} & \ding{51} & \texttt{\tiny ill.a.01/well.a.01} & .14 & .717 & .18 & 165 & \textsc{gn} & \ding{55} \\
\texttt{\tiny unfortunate.a.01/fortunate.a.01} & .50 & \textless.001 & 1.00 & 17 & \textsc{wp} & \ding{51} & \texttt{\tiny unfaithful.a.01/faithful.a.01} & .26 & .068 & .48 & 77 & \textsc{up} & \ding{55} \\
\texttt{\tiny segregated.a.01/integrated.a.03} & .48 & \textless.001 & 1.00 & 20 & \textsc{wp} & \ding{51} & \texttt{\tiny unambitious.a.01/ambitious.a.01} & .24 & .112 & .54 & 74 & \textsc{up} & \ding{51} \\
\texttt{\tiny illiterate.a.03/literate.a.02} & .48 & \textless.001 & 1.00 & 21 & \textsc{wp} & \ding{51} & \texttt{\tiny slow.a.01/fast.a.01} & .24 & .112 & .57 & 67 & \textsc{up} & \ding{55} \\
\texttt{\tiny straight.a.08/curved.a.01} & .48 & \textless.001 & .99 & 20 & \textsc{wp} & \ding{55} & \texttt{\tiny unfriendly.a.02/friendly.a.01} & .24 & .112 & .53 & 77 & \textsc{up} & \ding{51} \\
\texttt{\tiny ill-advised.a.01/well-advised.a.01} & .46 & \textless.001 & 1.00 & 21 & \textsc{wp} & \ding{51} & \texttt{\tiny unsettled.a.01/settled.a.01} & .24 & .112 & .45 & 92 & \textsc{up} & \ding{51} \\
\texttt{\tiny dull.a.02/bright.a.01} & .46 & \textless.001 & .99 & 24 & \textsc{wp} & \ding{51} & \texttt{\tiny harmless.a.01/harmful.a.01} & .22 & .179 & .43 & 94 & \textsc{up} & \ding{51} \\
\texttt{\tiny stupid.a.01/smart.a.01} & .46 & \textless.001 & 1.00 & 24 & \textsc{wp} & \ding{51} & \texttt{\tiny mild.a.01/intense.a.01} & .22 & .179 & .46 & 94 & \textsc{up} & \ding{51} \\
\texttt{\tiny destructive.a.01/constructive.a.01} & .46 & \textless.001 & 1.00 & 21 & \textsc{wp} & \ding{51} & \texttt{\tiny frivolous.a.01/serious.a.01} & .22 & .179 & .47 & 93 & \textsc{up} & \ding{51} \\
\texttt{\tiny pessimistic.a.01/optimistic.a.01} & .46 & \textless.001 & 1.00 & 24 & \textsc{wp} & \ding{55} & \texttt{\tiny foolish.a.01/wise.a.01} & .22 & .179 & .50 & 90 & \textsc{up} & \ding{55} \\
\texttt{\tiny tame.a.02/wild.a.01} & .46 & \textless.001 & 1.00 & 21 & \textsc{wp} & \ding{55} & \texttt{\tiny unworthy.a.01/worthy.a.01} & .22 & .179 & .53 & 81 & \textsc{up} & \ding{51} \\
\texttt{\tiny tasteless.a.02/tasteful.a.01} & .46 & \textless.001 & 1.00 & 24 & \textsc{wp} & \ding{55} & \texttt{\tiny unconventional.a.02/conventional.a.01} & .20 & .272 & .38 & 106 & \textsc{up} & \ding{55} \\
\texttt{\tiny intolerant.a.01/tolerant.a.01} & .46 & \textless.001 & 1.00 & 24 & \textsc{wp} & \ding{55} & \texttt{\tiny wrong.a.02/right.a.04} & .20 & .272 & .33 & 113 & \textsc{up} & \ding{51} \\
\texttt{\tiny powerless.a.01/powerful.a.01} & .44 & \textless.001 & 1.00 & 17 & \textsc{wp} & \ding{51} & \texttt{\tiny ungrateful.a.01/grateful.a.01} & .20 & .272 & .34 & 114 & \textsc{up} & \ding{51} \\
\texttt{\tiny unreasonable.a.01/reasonable.a.01} & .44 & \textless.001 & .99 & 21 & \textsc{wp} & \ding{51} & \texttt{\tiny negative.a.05/positive.a.08} & .20 & .272 & .33 & 113 & \textsc{up} & \ding{51} \\
\texttt{\tiny inferior.a.01/superior.a.02} & .42 & \textless.001 & .99 & 26 & \textsc{wp} & \ding{51} & \texttt{\tiny careless.a.01/careful.a.01} & .18 & .396 & .41 & 93 & \textsc{up} & \ding{55} \\
\texttt{\tiny unemployed.a.01/employed.a.01} & .40 & \textless.001 & .99 & 27 & \textsc{wp} & \ding{51} & \texttt{\tiny insane.a.01/sane.a.01} & .18 & .396 & .34 & 107 & \textsc{up} & \ding{55} \\
\texttt{\tiny retarded.a.01/precocious.a.01} & .40 & \textless.001 & .98 & 27 & \textsc{wp} & \ding{51} & \texttt{\tiny sweet.a.01/sour.a.02} & .64 & \textless.001 & 1.00 & 12 & \textsc{wp} & -- \\
\texttt{\tiny poor.a.02/rich.a.01} & .40 & \textless.001 & .95 & 35 & \textsc{wp} & \ding{55} & \texttt{\tiny green.a.03/ripe.a.01} & .46 & \textless.001 & 1.00 & 25 & \textsc{wp} & -- \\
\texttt{\tiny indirect.a.04/direct.a.03} & .38 & .001 & .96 & 31 & \textsc{wp} & \ding{51} & \texttt{\tiny short.a.06/long.a.05} & .40 & \textless.001 & .96 & 35 & \textsc{wp} & -- \\
\texttt{\tiny inhumane.a.01/humane.a.02} & .38 & .001 & .95 & 35 & \textsc{wp} & \ding{55} & \texttt{\tiny lost.a.03/saved.a.01} & .40 & \textless.001 & .99 & 25 & \textsc{wp} & -- \\
\texttt{\tiny quiet.a.02/noisy.a.01} & .38 & .001 & .94 & 37 & \textsc{wp} & \ding{55} & \texttt{\tiny intracellular.a.01/extracellular.a.01} & .40 & \textless.001 & .98 & 26 & \textsc{wp} & -- \\
\texttt{\tiny left.a.04/right.a.07} & .38 & .001 & .95 & 35 & \textsc{wp} & \ding{55} & \texttt{\tiny theoretical.a.02/applied.a.01} & .38 & .001 & .96 & 35 & \textsc{wp} & -- \\
\texttt{\tiny unsatisfactory.a.01/satisfactory.a.01} & .38 & .001 & .95 & 37 & \textsc{wp} & \ding{51} & \texttt{\tiny destroyed.a.01/preserved.a.02} & .36 & .003 & .92 & 37 & \textsc{wp} & -- \\
\texttt{\tiny stingy.a.01/generous.a.01} & .36 & .003 & .88 & 42 & \textsc{wp} & \ding{55} & \texttt{\tiny small.a.01/large.a.01} & .36 & .003 & .94 & 35 & \textsc{wp} & -- \\
\texttt{\tiny unimportant.a.01/important.a.01} & .36 & .003 & .97 & 35 & \textsc{wp} & \ding{55} & \texttt{\tiny rough.a.01/smooth.a.01} & .36 & .003 & .90 & 42 & \textsc{wp} & -- \\
\texttt{\tiny irregular.a.01/regular.a.01} & .36 & .003 & .96 & 35 & \textsc{wp} & \ding{51} & \texttt{\tiny acidic.a.01/alkaline.a.01} & .34 & .006 & .88 & 39 & \textsc{wp} & -- \\
\texttt{\tiny unwilling.a.01/willing.a.01} & .36 & .003 & .91 & 42 & \textsc{wp} & \ding{51} & \texttt{\tiny insufficient.a.01/sufficient.a.01} & .34 & .006 & .90 & 39 & \textsc{wp} & -- \\
\texttt{\tiny uncomfortable.a.01/comfortable.a.02} & .34 & .006 & .86 & 44 & \textsc{wp} & \ding{51} & \texttt{\tiny light.a.05/heavy.a.04} & .32 & .011 & .86 & 43 & \textsc{wp} & -- \\
\texttt{\tiny loose.a.01/compact.a.01} & .34 & .006 & .85 & 44 & \textsc{wp} & \ding{51} & \texttt{\tiny painless.a.02/painful.a.01} & .32 & .011 & .81 & 44 & \textsc{wp} & -- \\
\texttt{\tiny untrained.a.01/trained.a.01} & .34 & .006 & .89 & 42 & \textsc{wp} & \ding{51} & \texttt{\tiny unspecified.a.01/specified.a.01} & .28 & .039 & .83 & 48 & \textsc{wp} & -- \\
\texttt{\tiny insensitive.a.02/sensitive.a.02} & .34 & .006 & .94 & 36 & \textsc{wp} & \ding{51} & \texttt{\tiny dirty.a.01/clean.a.01} & .20 & .272 & .25 & 135 & \textsc{gn} & -- \\
\texttt{\tiny cowardly.a.01/brave.a.01} & .32 & .011 & .88 & 42 & \textsc{wp} & \ding{51} & \texttt{\tiny downtown.a.01/uptown.a.01} & .18 & .396 & .26 & 148 & \textsc{gn} & -- \\
\texttt{\tiny inconsiderate.a.01/considerate.a.01} & .32 & .011 & .90 & 42 & \textsc{wp} & \ding{51} & \texttt{\tiny asymmetrical.a.01/symmetrical.a.01} & .16 & .549 & .20 & 191 & \textsc{gn} & -- \\
\texttt{\tiny negligent.a.01/diligent.a.02} & .32 & .011 & .85 & 44 & \textsc{wp} & \ding{51} & \texttt{\tiny proximal.a.01/distal.a.01} & .12 & .869 & .14 & 211 & \textsc{gn} & -- \\
\texttt{\tiny lower-class.a.01/upper-class.a.01} & .30 & .022 & .86 & 44 & \textsc{wp} & \ding{51} & \texttt{\tiny horizontal.a.01/vertical.a.01} & .12 & .869 & .18 & 194 & \textsc{gn} & -- \\
\texttt{\tiny depressing.a.01/cheerful.a.01} & .30 & .022 & .84 & 50 & \textsc{wp} & \ding{51} & \texttt{\tiny earthly.a.01/heavenly.a.03} & .10 & .967 & .09 & 417 & \textsc{gn} & -- \\
\texttt{\tiny light-duty.a.01/heavy-duty.a.01} & .30 & .022 & .79 & 52 & \textsc{m} & \ding{55} & \texttt{\tiny central.a.02/peripheral.a.01} & .22 & .179 & .42 & 91 & \textsc{up} & -- \\
\texttt{\tiny dishonest.a.01/honest.a.01} & .30 & .022 & .77 & 51 & \textsc{m} & \ding{51} & \texttt{\tiny difficult.a.01/easy.a.01} & .22 & .179 & .50 & 84 & \textsc{up} & -- \\
\texttt{\tiny illogical.a.01/logical.a.01} & .30 & .022 & .77 & 52 & \textsc{m} & \ding{51} & \texttt{\tiny south.a.01/north.a.01} & .22 & .179 & .40 & 94 & \textsc{up} & -- \\
\texttt{\tiny following.a.03/leading.a.03} & .28 & .039 & .78 & 51 & \textsc{m} & \ding{51} & \texttt{\tiny worn.a.01/new.a.06} & .20 & .272 & .43 & 84 & \textsc{up} & -- \\
\bottomrule
\end{tabular}
\caption{Per-axis power analysis for \ours instantiated with \mistral across all 100 sampled semantic axes. $D$: two-sample Kolmogorov--Smirnov statistic (\secref{sec:hypothesis}); $p$: $p$-value; Pow.: empirical power at $n_{\mathcal{C}} = n_{\mathcal{C}'} = 50$ from 1{,}000 bootstrap resamples; $n_{80}$: estimated sample size per group to reach power $0.80$; Cat.: \textsc{wp} well-powered positive, \textsc{m} marginal, \textsc{gn} genuine null, \textsc{up} underpowered; Hum.: human annotators judge the semantic axis stereotypical (\ding{51}), non-stereotypical (\ding{55}), or not applicable (--).}
\label{tab:power_mistral_full_100}
\end{table*}

\section{Code and Computational Resources}
The code and data are provided as supplementary material 
and will be made publicly available with the camera-ready 
version of this paper. 
All experiments were conducted on NVIDIA A100 40GB GPUs. 
Across all experiments, including ablations, the total compute budget was approximately 10 GPU-hours per model for \llama and \mistral.

\section{Licenses and AI Tool Use}
\paragraph{Licenses.} WordNet is distributed under Princeton University's WordNet license agreement, \llama under the Meta Llama 3 Community License, and \mistral under Apache 2.0; both LLMs were obtained from Hugging Face~\footnote{\url{https://huggingface.co/meta-llama/Meta-Llama-3-8B-Instruct}}$^{,}$\footnote{\url{https://huggingface.co/mistralai/Mistral-7B-Instruct-v0.1}}. \texttt{wordfreq} and \texttt{SciPy} are open-source. All resources are used solely for research purposes and in accordance with their respective licenses and terms of use.

\paragraph{Use of AI Assistants.} Generative AI tools were used to assist with code development (e.g., code completion) and to refine the clarity and style of the manuscript text.

\end{document}